\documentclass[runningheads]{llncs}

\usepackage{eccv}

\usepackage{eccvabbrv}

\usepackage{pgfplots}
\usepackage{graphicx}
\usepackage{amsmath}
\usepackage{wrapfig}
\usepackage{multirow} 
\usepackage{amssymb}
\usepackage{booktabs}
\usepackage{color, colortbl}
\usepackage{xspace}
\usepackage{tablefootnote}
\usepackage{xparse}
\usepackage{dsfont}

\definecolor{LightCyan}{rgb}{0.88,1,1}

\usepackage[accsupp]{axessibility}  

\usepackage[pagebackref,breaklinks,colorlinks,citecolor=eccvblue]{hyperref}

\usepackage{orcidlink}

\begin{document}

\title{Elevating \textit{All} Zero-Shot Sketch-Based Image Retrieval Through Multimodal Prompt Learning} 

\titlerunning{\textsc{SpLIP}}


\author{Mainak Singha\inst{1}\orcidlink{0000-0002-7615-2575} \and
Ankit Jha\inst{1,2}\orcidlink{0000-0002-1063-8978} \and
Divyam Gupta\inst{1}\orcidlink{0009-0006-7294-3739}\and
Pranav Singla\inst{1}\orcidlink{0009-0004-3812-8517}\and
Biplab Banerjee\inst{1}\orcidlink{0000-0001-8371-8138}}

\authorrunning{Singha et al.}

\institute{Indian Institute of Technology Bombay, India \and
INRIA, Grenoble, France\\
\email{\{mainaksingha.iitb, ankitjha16, divsg1803, pranavsingla.cminds.iitb, getbiplab\}@gmail.com}}

\maketitle

\begin{abstract}
We address the challenges inherent in sketch-based image retrieval (SBIR) across various settings, including zero-shot SBIR, generalized zero-shot SBIR, and fine-grained zero-shot SBIR, by leveraging the vision-language foundation model CLIP. While recent endeavors have employed CLIP to enhance SBIR, these approaches predominantly follow uni-modal prompt processing and overlook to exploit CLIP's integrated visual and textual capabilities fully. To bridge this gap, we introduce \textsc{SpLIP}, a novel multi-modal prompt learning scheme designed to operate effectively with frozen CLIP backbones. We diverge from existing multi-modal prompting methods that treat visual and textual prompts independently or integrate them in a limited fashion, leading to suboptimal generalization.
\textsc{SpLIP} implements a bi-directional prompt-sharing strategy that enables mutual knowledge exchange between CLIP's visual and textual encoders, fostering a more cohesive and synergistic prompt processing mechanism that significantly reduces the semantic gap between the sketch and photo embeddings. In addition to pioneering multi-modal prompt learning, we propose two innovative strategies for further refining the embedding space. The first is an adaptive margin generation for the sketch-photo triplet loss, regulated by CLIP's class textual embeddings. The second introduces a novel task, termed conditional cross-modal jigsaw, aimed at enhancing fine-grained sketch-photo alignment by implicitly modeling sketches' viable patch arrangement using knowledge of unshuffled photos.
Our comprehensive experimental evaluations across multiple benchmarks demonstrate the superior performance of \textsc{SpLIP} in all three SBIR scenarios. Project page: \url{https://mainaksingha01.github.io/SpLIP/}.

\end{abstract}

\begin{figure}[ht!]
    \centering
    \includegraphics[width=\textwidth]{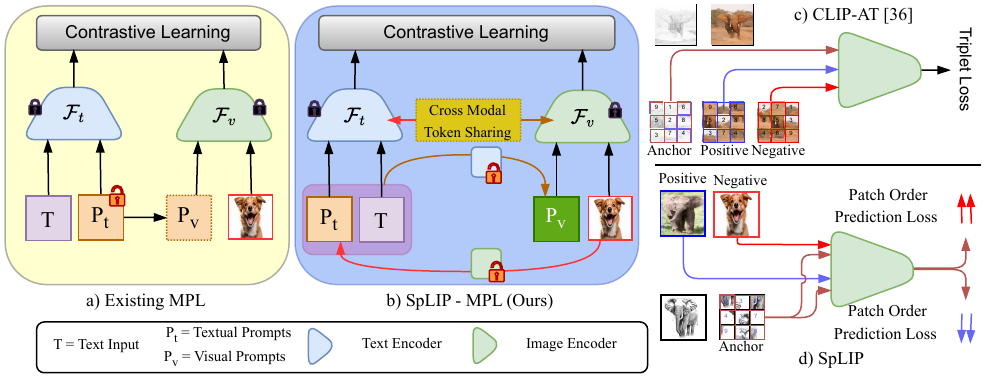}
    \vspace{-0.7cm}
    \caption{\textbf{(a, b) The difference between the existing multi-modal prompt learning (MPL) vs ours}. As opposed to the literature \cite{maple, promptsrc, coprompt}, we propose to enhance the generalizability of both the textual and visual prompt embeddings with mutual knowledge sharing in a principled layer-wise fashion. \textbf{(c, d) Proposed conditional cross-modal jigsaw vs the literature \cite{clipforall}}. As against a triplet loss connecting sketch-photo with the same permutation while contrasting against the photo with a different permutation, we propose better learning of patch arrangements by positively associating a permuted sketch with its intact photo counterpart through a novel objective. }
    \label{fig:teaser}
    \vspace{-0.7cm}
\end{figure}

\section{Introduction}
\label{sec:intro}
Hand-drawn sketches adeptly convey abstract ideas with their simple yet evocative lines. The advent of touchscreen mobile devices has propelled sketch-based image retrieval (SBIR) \cite{survey1, survey2} into the limelight, offering myriad practical uses. SBIR retrieves photos from a vast repository based on the same category as a query sketch. However, despite representing the same class, sketches and photos often differ noticeably due to their distinct domains. Traditional methods \cite{eitz2010sketch, stylemeup} address domain heterogeneity, assuming full visibility of test classes during training, showing promise in effective retrieval. Yet, a more realistic challenge emerges when test set categories remain unseen during training, termed zero-shot SBIR (ZS-SBIR) \cite{sempcyc, bda, sake}. Contrarily, generalized ZS-SBIR (GZS-SBIR) \cite{sempcyc} encompasses retrieval photos of known and novel classes for novel-class sketch queries during inference, heightening complexity. Additionally, instance-level fine-grained ZS-SBIR (FG-ZS-SBIR) \cite{sketchmeshoe, ccdg} focuses on precise shape matching, intensifying challenges compared to category-level SBIR.

The crux of all the SBIR settings lies in learning an embedding space for the sketches and photos that is \textit{unbiased} to the training data, \textit{discriminative} given data from both the photo and sketch modalities and hence, \textit{domain-agnostic}.

Leading SBIR frameworks leveraging ConvNets and ViTs \cite{bda, sempcyc, tvt, pskd} face intrinsic semantic constraints by their architecture. On the contrary, the advent of multi-modal foundational models, notably CLIP \cite{clip} and Align \cite{align}, has markedly enhanced visual comprehension by integrating visual and textual information. These VLMs have excelled in cross-domain learning, but their SBIR integration remains limited.
Initiatives like \cite{clipforall} have aimed to tailor CLIP to ZS-SBIR and FG-ZS-SBIR, focusing on visual prompts and patch shuffling to align sketches and photos at both micro and macro levels. Other efforts \cite{marl, tlt} seek to finetune CLIP's embeddings for SBIR, indicating an increasing desire to leverage these multi-modal platforms beyond original purposes.

\noindent \textbf{Highlighting research gaps:} Despite the success, these strategies often rely on simplistic, one-dimensional prompting, failing to harness CLIP's dual-pathway synergy fully. This overlooks CLIP's visual-textual fusion's rich, complementary insights, leading to suboptimal ZS-SBIR performance. Addressing this, there's an urgent call for novel approaches that dynamically utilize this combined knowledge to surpass the semantic limitations of existing models, thus expanding SBIR potentials.
  
Moving away from singular prompting, multi-modal methods recommended by \cite{maple, coprompt, promptsrc} employ simultaneous prompting across CLIP's pathways, thus narrowing the semantic gap in embeddings. Nonetheless, these approaches struggle with one-sided prompt integration and the dismissal of static textual elements, causing a reduction in semantic depth. Particularly, the adaptability of the textual pathway is compromised, remaining insensitive to visual nuances, even as \cite{cocoop} underscores the importance of enhancing textual adaptability in CLIP. We advocate for a more cohesive and dynamic knowledge interchange between the visual and textual domains to enhance the flexibility of (G)ZS-SBIR. Furthermore, although the patch shuffling strategy by \cite{clipforall} enhances the shape equivalence of images and sketches, especially for FG-ZS-SBIR, indiscriminate matching of patch-permuted sketch-photo versions without regard to the natural qualities of objects could lead to overfitting. We recommend matching sketch patch permutations with semantically corresponding, unshuffled photos and the reverse, helping embeddings to discern how patch arrangements correspond with the entirety of an image's objects.

\noindent \textbf{Proposed solution:} Our proposed model, \textsc{SpLIP}, tackles these challenges by implementing a novel deep multimodal prompting approach (Fig. \ref{fig:teaser}), facilitating efficient knowledge exchange between text and image branches of frozen CLIP. In CLIP's text encoder, we enhance static textual tokens by incorporating additional latent visual tokens at each layer, departing from previous random initialization methods \cite{maple, coprompt}. Likewise, we enrich the image embeddings within the CLIP vision backbone with information from these image-conditioned textual token embeddings, summarized layer by layer over all the semantic categories. This \textit{bidirectional} information sharing mitigates the semantic gap in the obtained embeddings, thus contributing positively towards zero-shot inference.

As we advance, we introduce a novel conditional cross-modal jigsaw task to strengthen the linkage between photo and sketch pairs for all the considered SBIR variants, predominantly for FG-ZS-SBIR. This method requires solving a jigsaw puzzle \cite{noroozi2016unsupervised} using an anchor sketch, assisted by a positive and a negative image from the alternate modality while ensuring that the positive image significantly aids in this process, thereby enhancing model generalization through the learning of patch arrangement for reconstructing a complete image. This approach deviates from previous works \cite{clipforall, pang2020solving} that either deal with mixed-modal images for a naive jigsaw solver or employ triplet objectives with uniform or varying patch permutations for identifying positive and negative pairs, thus failing to bridge the gap between local and global contexts effectively (Fig. \ref{fig:teaser}).
 Finally, we integrate the gold-standard cross-modal triplet loss for SBIR and a sketch/photo-text-based classification loss of CLIP, introducing an adaptive margin scheme for the triplet objective derived from the semantic class-prompt embeddings.
Our salient contributions are, therefore,

   
 \textbf{-} Introducing \textsc{SpLIP}, a novel deep multi-modal prompt tuning model within the realm of CLIP tailored for ZS-SBIR and FG-ZS-SBIR tasks. It introduces a more systematic cross-modal mutual guidance within CLIP's text and vision encoders. To the best of knowledge, this is the first endeavor of multi-modal prompting for solving ZS-SBIR variants.

 \textbf{-} We enhance the conventional cross-modal triplet loss objective in ZS-SBIR by incorporating an adaptive margin scheme, leveraging CLIP's textual prompt embeddings. Additionally, we introduce a novel conditional cross-modal jigsaw task aimed at refining the association between sketch and photo pairs.

 \textbf{-} We conduct extensive experiments on three benchmark datasets: Sketchy-Ext \cite{sketchy, sempcyc, caae}, TU-Berlin-Ext \cite{tuberlin, tuberlinextend}, and QuickDraw-Ext \cite{quickdraw, doodle}, covering (G)ZS-SBIR and FG-ZS-SBIR settings. \textsc{SpLIP} consistently outperforms existing competitors, achieving significant improvements in all the metrics.

    

\vspace{-0.1cm}
\section{Related Works}
\vspace{-0.3cm}
\subsection{Sketch-based image retrieval (SBIR)}
\vspace{-0.2cm}
\noindent\textbf{(Generalized) Zero-shot SBIR:} SBIR tasks involve retrieving photos corresponding to specific categories from a diverse collection of multi-category images based on a given sketch query, demanding a thorough understanding of the joint sketch-photo manifold. The literature is rich in fully-supervised SBIR endeavors utilizing deep and hand-crafted descriptors and involving different learning mechanisms, including both generative and discriminative approaches \cite{sketchy,hu2013performance, zhou2012sketch, deepspatial,crossdomain,stylemeup,fgsbir}. Recently, \cite{chaudhuri2023data} introduced a data-free training strategy for SBIR, relaxing the need for curating sketch-photo pairs.

Researchers have addressed challenges in recognizing unseen test-time classes through ZS-SBIR, facilitating category-level generalization. Extending the traditional ConvNet-based frameworks \cite{doodle, sempcyc, caae, sake}, graph convolutional networks, ViTs, and their combinations have been introduced to learn an unbiased shared feature space \cite{zse, gupta2022zero, zhang2020zero}. Recent advancements \cite{clipforall, tlt, marl, adapt} leverage CLIP's zero-shot inference by integrating text with sketches and photos, outperforming counterparts. A combined loss function of supervised cross-entropy and metric objectives fosters a discriminative embedding space. On the other hand, GZS-SBIR permits the presence of training and test time photos during inference, causing the model to show high bias towards the training data, which has been tackled in the literature from different perspectives \cite{sempcyc, bda, ocean, saa, zse, stl, marl}. \textit{We take a different route to tackle (G/FG)ZS-SBIR through multi-modal prompting in CLIP, thus modeling the visual-semantic synergy effectively.}

\noindent\textbf{Fine-grained ZS-SBIR:} Transitioning from categorical ZS-SBIR, FG-ZS-SBIR aims at identifying specific photos relating to sketches at an instance level. Initially rooted in a deep triplet-ranking Siamese framework \cite{sketchmeshoe}, FG-ZS-SBIR's evolution incorporates attention modules \cite{deepspatial, crosshier}, hybrid cross-domain mapping \cite{crossdomain}, and manifold modeling for universality \cite{adaptivefg}. Enhancements proceed with an intra-modal triplet goal, solutions for sparse sketch annotations \cite{sketchpvt}, patch similarity via cross-interaction \cite{sun2022dli}, and innovative patch shuffling \cite{clipforall, pang2020solving}. \textit{Exploiting patch shuffling for precise shape alignment and utilizing CLIP's robust capabilities, we propose a unique conditional cross-modal jigsaw challenge. Aimed at refining alignment between permutations of sketch patches and intact photos, this initiative significantly deepens contextual comprehension from a local to a global scale, marking a pivotal advance in
FG-ZS-SBIR development.}

\vspace{-3mm}
\subsection{Vision-language models and multi-modal prompt learning}
Vision-Language Models (VLMs) like CLIP \cite{clip} and VisualBERT \cite{visualbert} have revolutionized computer vision by merging visual and textual data through multimodal learning, creating detailed representations. Leveraging textual features from language models (e.g., BERT \cite{bert}, GPT \cite{gpt}) and visual inputs from ConvNets or ViTs \cite{vit}, these models achieve semantic depth and exhibit strong zero-shot inference for varied tasks \cite{maple, stylip}.

Recent research \cite{coop, cocoop, stylip, lasp, odgclip, gopro, adclip} highlights prompt learning as a viable alternative to VLM fine-tuning on downstream tasks. To this end, unlike uni-modal approaches for CLIP, multi-modal deep prompts synergize its visual and textual components. \cite{maple} proposed to learn both visual and textual prompts and showed that initializing the visual prompts from the textual counterparts enhances the performance. In addition, \cite{promptsrc} and \cite{coprompt} introduced regularization and feature consistency to prevent overfitting and ensure textual variety. \textit{However, existing methods mainly employ unidirectional token sharing, restricting the overall generalizability and semantic depth of the learned embeddings. Contrarily, we propose a bilateral approach to disseminate relevant cross-modal insights across CLIP’s branches, establishing a more effective multi-modal prompting paradigm.}
\begin{figure*}[ht!]
    \centering
    \vspace{-.2cm}
    \includegraphics[width=\textwidth]{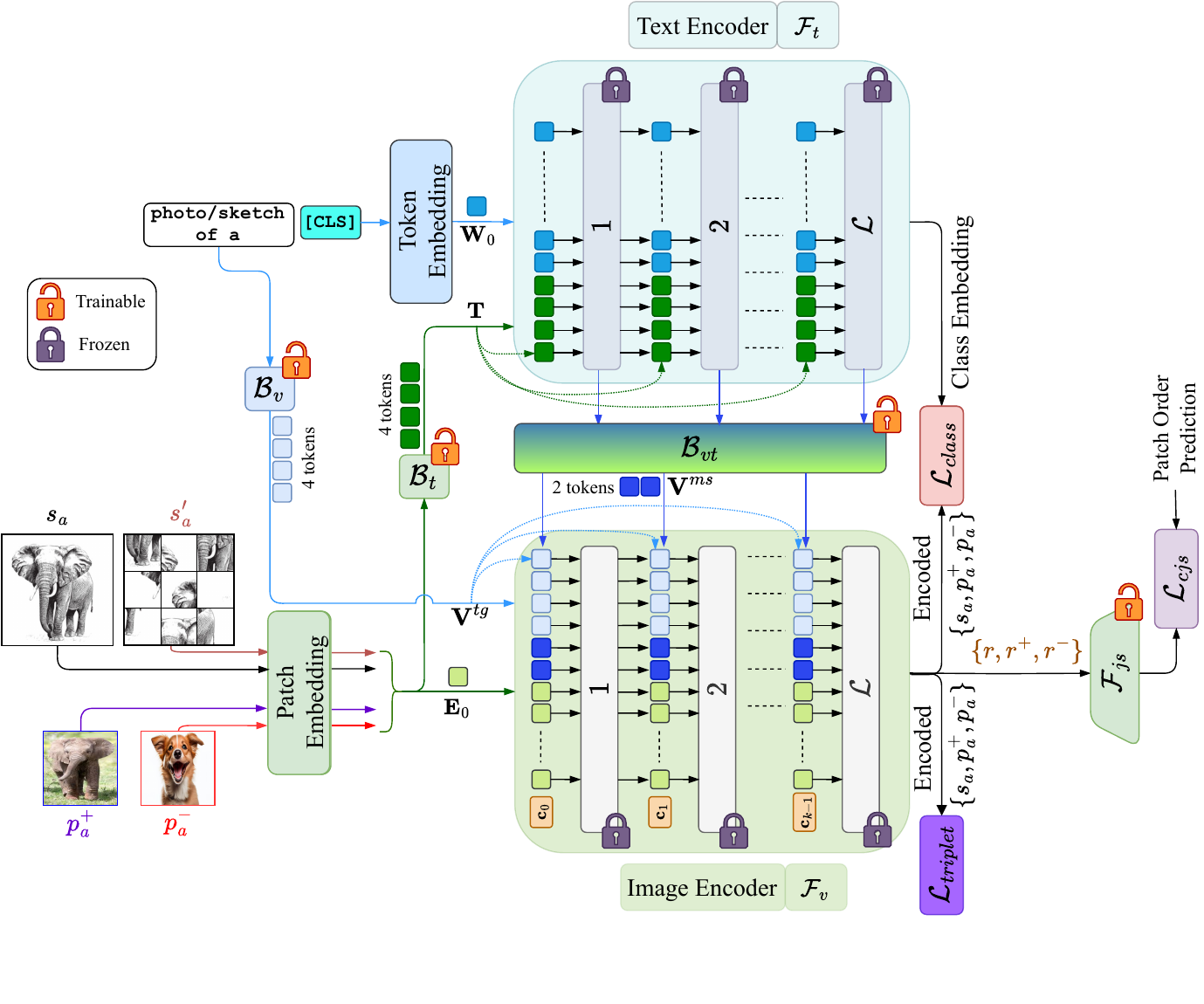}
    \vspace{-1.5cm}
    \caption{\small{\textbf{The model architecture for \textsc{SpLIP}}, which capitalizes on CLIP's static text and vision backbones, $\mathcal{F}_t$ and $\mathcal{F}_v$, introducing a bidirectional prompt exchange. Image patch embeddings, transformed via $\mathcal{B}_t$, generate textual tokens $\mathbf{T}$ for different layers of $\mathcal{F}_t$. Similarly, $\mathcal{F}_v$ layers receive "\texttt{sketch/photo of a}" token embeddings $\mathbf{V}^{\texttt{tg}}$ through $\mathcal{B}_v$ and consolidated prompt tokens from $\mathcal{F}_t^{l}$, across all semantic classes in $\mathcal{C}^s$, $\mathbf{V}^{\texttt{ms}}$, via $\mathcal{B}_{vt}$. This setup enriches both textual and visual pathways with diverse information sources. The model also tackles a unique conditional cross-modal jigsaw task, with a decoder $\mathcal{F}_{js}$ processing sketch-photo pairs from $(s_a, s'_a, p_a^+, p_a^-)$ to understand complex relationships through pairwise fused features of $s'_a$  and the remaining counterparts, $(r, r^+, r^-)$. Training involves a blend of loss functions: photo-sketch triplet loss $\mathcal{F}_{triplet}$, text-image classification loss $\mathcal{F}_{class}$, and proposed jigsaw loss $\mathcal{L}_{cjs}$. During inference on $\mathcal{G}^u$, $\mathbf{V}^{\texttt{ms}}$ is derived leveraging the classes in $\mathcal{C}^s$, overlooking the need of $\mathcal{C}^u$, and leading to a nearest-neighbor based ranking of photos for sketch queries in the output of $\mathcal{F}_v$.}}
    \label{architecture}
    \vspace{-0.4cm}
\end{figure*}

\vspace{-3mm}
\section{Methodology}\vspace{-2mm}
SBIR entails retrieving $\mathcal{K}$ photos $\{p_k\}_{k=1}^{\mathcal{K}} \in \mathcal{P}$ from a gallery ($\mathcal{G}$), given a query-sketch ($s \in \mathcal{S}$) belonging to a specific category out of a total of $\mathcal{C}$ classes. In zero-shot tasks, $\mathcal{C}$ is divided into seen training classes ($\mathcal{C}^{s}$) and unseen testing classes ($\mathcal{C}^{u}$), where $\mathcal{C} = \mathcal{C}^{s} \cup \mathcal{C}^{u}$ and $\mathcal{C}^{s} \cap \mathcal{C}^{u} = \emptyset$.

The training dataset $\mathcal{G}^{s} = (\mathcal{S}^{s}, \mathcal{P}^{s}, \mathcal{C}^{s})$ comprises sketches $\mathcal{S}^{s}$ and photos $\mathcal{P}^{s}$ from $\mathcal{C}^{s}$ categories. During inference, the gallery $\mathcal{G}^{u} = (\mathcal{S}^{u}, \mathcal{P}^{u}, \mathcal{C}^{u})$ containing sketches $\mathcal{S}^u$ and photos $\mathcal{P}^u$ with category labels in $\mathcal{C}^u$ is involved. In contrast to ZS-SBIR, the GZS-SBIR setup considers photos in $\mathcal{P}^s \cup \mathcal{P}^u$ for a given sketch query $s^u \in \mathcal{S}^u$ for retrieval during inference. FG-ZS-SBIR  aims at instance-level sketch-photo matching within specific categories \cite{sketchmeshoe}, in contrast to the broader category-level focus of conventional SBIR methods. Following \cite{clipforall}, we consider the multi-category FG-ZS-SBIR setting where paired sketch-photo instances are available from multiple categories. 

In the following, we delve into the initialization of text inputs and provide a detailed explanation of the image-text embeddings of CLIP in Section \ref{multiclip}. Moving forward, our image-driven textual prompting approach is elaborated upon in Section \ref{textualprompt}, while Section \ref{visualprompt} addresses the proposed visual prompting methodology. For fine-grained sketch-photo feature association, we discuss the proposed conditional cross-modality jigsaw task and the related details in Section \ref{decoding}. Finally, in Section \ref{loss}, we mention the considered loss objectives. A list of important variables is summarized in the \textbf{Supplementary}.

\vspace{-0.1cm}
\subsection{Initialization of visual-textual embeddings} 
\label{multiclip} 
\vspace{-0.1cm}
The pre-trained CLIP model operates on two modalities: text and image. It consists of a transformer-based \cite{transformer} text encoder ($\mathcal{F}_t$) and ViT-based \cite{vit} image encoder ($\mathcal{F}_v$). They both contain $\mathcal{L}$ transformer encoder layers. $\mathcal{F}_t$ generates feature representations for text descriptions to capture semantic information. Initially, it tokenizes the text inputs, consisting of $\mathcal{J}$ words, and projects them into word embeddings $\mathbf{W}_{0} = [w^{1}_{0}, w^{2}_{0}, \cdots, w^{\mathcal{M}}_{0}] \in \mathbb{R}^{\mathcal{M} \times d_t}$, where $[.,.]$ denotes stacking and concatenation, $\mathcal{M}$ refers the number of embedding tokens ($77$ per class-prompt) and $d_t$ is the dimension of text tokens. In our approach, we use the input texts as \texttt{``sketch/photo of a [CLS]''} for sketches and photos, respectively, for (G)ZS-SBIR, meaning $\mathcal{J}=5$. Here, the \texttt{[CLS]} token represents class embeddings, completing the prompt embedding $\mathbf{W}_{0}$. For the FG-ZS-SBIR task, we use common text input, \texttt{``visual representation of [CLS]''}, for both sketches and photos. We obtain the $l$-th layer embedding of $\mathcal{F}_t$, denoted as $\mathcal{F}_t^l$, as follows,\vspace{-.2cm}
\begin{equation}
[\mathbf{W}_{l}] = \mathcal{F}_t^l(\mathbf{W}_{l-1}) \in \mathbb{R}^{\mathcal{M} \times d_t} \qquad l=1, 2, \cdots, \mathcal{L}
\end{equation}

On the image side, the images from $\mathcal{P}$ and $\mathcal{S}$ are partitioned into fixed-size patches and encoded through $\mathcal{F}_v$. Each patch undergoes projection to generate initial patch embeddings ($\mathbf{E}_{0}$), along with a learnable \texttt{cls} token $\mathbf{c}_0$, denoted as $[\mathbf{c}_{0}, \mathbf{E}_{0}] \in \mathbb{R}^{1 + \mathcal{N} \times d_v}$, where $\mathcal{N}$ denotes the number of patches and $d_v$ is the dimension of patch tokens. Henceforth, the output embedded tokens of the $l$-th layer of $\mathcal{F}_v$ can be expressed as,\vspace{-.2cm}
\begin{equation}
[\mathbf{c}_l,  \mathbf{E}_l] = \mathcal{F}_v^l([\mathbf{c}_{l-1}, \mathbf{E}_{l-1}]) \in \mathbb{R}^{1 + \mathcal{N} \times d_v}  \qquad l=1, 2, \cdots, \mathcal{L}
\vspace{-0.1cm}
\end{equation}

\subsection{Proposed vision-guided deep textual prompting}
\label{textualprompt}
\vspace{-0.1cm}
We propose a novel \textit{vision-guided deep textual prompting} approach, where deep prompting determines slightly changing the input raw tokens of each of the layers of $\mathcal{F}_t$ for both text inputs associated with sketches and photos. Our proposal involves incorporating visual information into the tuning process of textual prompts. Specifically, we introduce a \textit{visual-to-textual mapping} block ($\mathcal{B}_t$), which generates \textit{m} learnable prompt tokens, denoted ($\mathcal{T}_{1:m}$) collectively as $(\mathbf{T})$, from the $\mathcal{N}$ visual patch embeddings $\mathbf{E}_{0}$ (Fig. \ref{architecture}).

Precisely, each layer in $\mathcal{F}_t$ receives ($\mathbf{T}$) in the corresponding input space.  In the first layer, $\mathbf{T}$ (aka $\mathbf{T}_0$) replaces $m$ tokens of $\mathbf{W}_0$, and the input embedding for the first layer of $\mathcal{F}_{t}$ becomes $[\mathbf{T}_{0}; \mathbf{W}_{0}]$. Here, {$[a ; b]$ denotes stacking after replacing a similar number of tokens of $b$ with all of the tokens of $a$}. Consequently, for the $l$-th layer, $\mathbf{T}_{l} = \mathbf{T} = \mathcal{B}_{t}(\mathbf{E}_0)$.

As already pointed out, our approach differs from existing literature \cite{maple, coprompt} in that our learnable tokens in the textual prompts capture visual distributions, as opposed to the random initialization approach followed by \cite{maple, coprompt}.

Finally, the output operation for the $l$-th layer can be expressed as,
\vspace{-.2cm}
\begin{equation}
[\_,  \mathbf{W}_{l}] = \mathcal{F}_t^l([\mathbf{T}_{l-1} ; \mathbf{W}_{l-1}]) \in \mathbb{R}^{\mathcal{M} \times d_t} \qquad l=1, 2, \cdots, \mathcal{L}
\vspace{-0.1cm}
\end{equation}

\subsection{Proposed text-guided deep visual prompting} 
\label{visualprompt}
\vspace{-0.1cm}
Our visual prompting approach leverages textual information within CLIP via two distinct mechanisms. We harness the initial tokenized text input excluding the [CLS] token, denoted as $\mathcal{W'}$ comprising of $(\mathcal{J}-1)$ tokens. These are employed as \textit{semantic domain knowledge} (${\mathbf{V}^{\texttt{tg}}}$), which is then mirrored by an equivalent stack of $(\mathcal{J}-1)$ learnable tokens ($\mathcal{V}^{\texttt{tg}}_{1:\mathcal{J}-1}$). This mirroring is facilitated via a \textit{textual-to-visual mapping block} ($\mathcal{B}_v$), which operates across all layers of $\mathcal{F}_v$. For any given layer $(l+1)$ within $\mathcal{F}_v$, this process can be succinctly described as: $\mathbf{V}^{\texttt{tg}}_{l} = \mathbf{V}^{\texttt{tg}} = \mathcal{B}_{v}(\mathcal{W'})$, effectively embedding semantic textual insights into the visual domain for enhanced model comprehension and interaction (Fig. \ref{architecture}).


In addition, we enforce token sharing from each layer of $\mathcal{F}_t$ to the respective layer input of $\mathcal{F}_v$. Note that, from $l \geq 1$, $\mathbf{W}_l$ implicitly includes visual knowledge as per our proposed textual prompting.
Secondly, as opposed to the visual prompting approach proposed in \cite{maple}, which only shares the learnable textual tokens with the visual branch, we propose to consider all the tokens from $\mathbf{W}_l$ over all the classes present in $\mathcal{C}^s$ to be included into the input space of $\mathcal{F}_v^l$. This class-agnostic knowledge-sharing approach helps diminish the semantic gap in the learned embeddings.

We proceed by transferring the output ($\mathbf{W}_{l}$) of the $l$-th layer of $\mathcal{F}_{t}$ to a \textit{vision-text conjunction block} ($\mathcal{B}_{vt}$), which then generates scale-specific inputs ($\mathbf{V}^{\texttt{ms}}_{l-1}$) for the corresponding $l$-th layer of $\mathcal{F}_{v}$ consisting of $n$ learnable tokens ($\mathcal{V}^{\texttt{ms}}_{1:n}$). As mentioned, $\mathcal{B}_{vt}$ takes all of the class-defined text feature tokens of $\mathbf{W}_{l}$, as input. Evidently, unlike $\mathbf{V}^{\texttt{tg}}$, the prompt tokens of $\mathbf{V}^{\texttt{ms}}$ are not similar to each other for every layers. We note that ($\mathcal{B}_{vt}$) is shared across all the encoder layers of $\mathcal{F}_t$. 
For the $l$-th layer of $\mathcal{F}_{v}$, $\mathbf{V}^{\texttt{ms}}$ can be defined as,
\vspace{-0.2cm}
\begin{equation}
\mathbf{V}^{\texttt{ms}}_{l-1} = \{\mathcal{V}^{\texttt{ms}}_{k_{l-1}} \in \mathbb{R}^{d_v}\}^{n}_{k=1} = \mathcal{B}_{vt}(\mathcal{F}_t^l([\mathbf{T}_{l-1} ; \mathbf{W}_{l-1}])) \in \mathbb{R}^{n \times d_v},   \qquad l=1, 2, \cdots, \mathcal{L}
\end{equation}

Finally, we concat the generated prompt tokens of $\mathbf{V}^{\texttt{tg}}$ and $\mathbf{V}^{\texttt{ms}}$ for each of the layers, expressed as a common \textit{visual prompt} ($\mathbf{V}$) i.e. for inputting to the $l$-th layer of $\mathcal{F}_{v}$: $\mathbf{V}_{l-1} = [ \mathbf{V}^{\texttt{tg}}_{l-1} , \mathbf{V}^{\texttt{ms}}_{l-1}]$. Hence, the processing at the $l$-th layer of $\mathcal{F}_v$ is mentioned as,
\vspace{-0.2cm}
\begin{equation}
[\mathbf{c}_l , \_ ; \mathbf{E}_l] = \mathcal{F}_v^l([\mathbf{c}_{l-1} , \mathbf{V}_{l-1} ; \mathbf{E}_{l-1}]) \in \mathbb{R}^{1 + \mathcal{N} \times d_v}  \qquad l=1,2,\cdots,\mathcal{L}
\vspace{-0.1cm}
\end{equation}

\subsection{Proposed conditional cross-modal jigsaw solver for fine-grained sketch-photo feature association}
\label{decoding}
\vspace{-0.1cm}

Furthermore, we introduce the task of \textit{conditional cross-modal jigsaw} to enhance the intricate connections between photos and sketches belonging to identical classes (or specific instances in the context of FG-ZS-SBIR). This technique marks a departure from previous methods, such as the one by \cite{pang2020solving}, which created a hybrid image by interspersing patches between sketches and photos randomly and pre-training the feature extraction backbone to predict the sequence of patches in this blended image. Similarly, \cite{clipforall} utilized a uniform permutation for sketch-photo pairs to delineate positive pairs, contrasting this with a distinct permutation on the photo to forge the negative pair. However, these methods grapple with the challenge of precisely aligning the shuffled image with its original format, a critical step for accurately learning patch arrangements in detail and maintaining the spatial coherence within the images.

To address these challenges, our proposed method incorporates positive and negative counterparts ($p_a^+, p_a^-$) from set $\mathcal{P}^s$ when resolving the jigsaw puzzle for a sample $s_a \in \mathcal{S}^s$ given its permuted version $s_a'$ which is obtained through a permutation function $\delta$, given the random permutation $y^{\text{perm}} \in \mathcal{Y}^{\text{perm}}$: $s'_a = \delta (s_a, y^{\text{perm}})$. Here, the information embedded in $(s'_a, p_a^+)$ and $(s'_a, p_a^-)$ is then processed through a transformer-based jigsaw-solver network, $\mathcal{F}_{js}$, operating atop $\mathcal{F}_v$, to leverage $p_a^+$ to inform better the prediction of the permutation arrangements of $s'_a$ compared to $p_a^-$, achieved through a hinge objective. Collectively, $\mathcal{F}_{js}$ directly resolves the jigsaw puzzle for $s'_a$ when coupled with $s_a$. These combined losses are called $\mathcal{L}_{cjs}$.
For simplicity, we define the fused inputs from pairs of images to $\mathcal{F}_{js}$ as follows: $r = [\mathcal{F}_v(s_a),\mathcal{F}_v(s'_a)]$, $r^+ = [\mathcal{F}_v(p_a^+),\mathcal{F}_v(s'_a)]$, and $r^- = [\mathcal{F}_v(p_a^-),\mathcal{F}_v(s'_a)]$, respectively.

\vspace{-0.1cm}
\subsection{Loss functions and inference}
\label{loss}
\vspace{-0.1cm}
Following \cite{clipforall}, we train the LayerNorm parameters $({\theta}, {\phi})$ of $\mathcal{F}_v$ and $\mathcal{F}_t$, together with $(\mathcal{B}_v, \mathcal{B}_t, \mathcal{B}_{vt}, \mathcal{F}_{js})$ while keeping the other layers fixed.
\vspace{1mm}

\noindent $\mathcal{L}_{triplet}$: \textbf{Cross-visual modality triplet loss with proposed adaptive margin:} Given the triplet of embeddings $(\mathcal{F}_v(s_a), \mathcal{F}_v(p_a^+), \mathcal{F}_v(p_a^-))$, we formulate a triplet objective aimed at minimizing the distance between $\mathcal{F}_v(s_a)$ and $\mathcal{F}_v(p_a^+)$ while maximizing the distance between $\mathcal{F}_v(s_a)$ and $\mathcal{F}_v(p_a^-)$. Traditional triplet objectives employ a fixed margin, which may not be optimal for zero-shot tasks. In contrast, we propose leveraging the semantic space of CLIP to define the margin, utilizing the embeddings of positive and negative class names from $\mathcal{F}_t$. Let $\mathcal{F}_t(c^+)$ and $\mathcal{F}_t(c^-)$ represent the semantic class embeddings for $p_a^+/s_a$ and $p_a^-$, respectively. We define $\mu(c^+, c^-) =\cos(\mathcal{F}_t(c^+),\mathcal{F}_t(c^-))$. Consequently, the margin increases when $c^+$ and $c^-$ are semantically close. The loss is,
\vspace{-0.1cm}
\scriptsize
\begin{equation}
\vspace{-0.1cm}
\mathcal{L}_{\text{triplet}} = \underset{\begin{array}{c}
    \mathcal{B}_{v}, \mathcal{B}_{t}, \mathcal{B}_{vt}, \\
    {\theta}, {\phi}
    \end{array}}{\min} \sum_{(s_a, p_a^+, p_a^-) \in \mathcal{G}^s} \left[ \|\mathcal{F}_v(s_a) - \mathcal{F}_v(p_a^+)\|_2^2 - \|\mathcal{F}_v(s_a) - \mathcal{F}_v(p_a^-)\|_2^2 + \mu(c^+, c^-) \right]_+
\end{equation}
\normalsize

\noindent \textbf{$\mathcal{L}_{class}$: Text-image classification loss:} A supervised contrastive loss is introduced to correctly classify sketches and photos in $\mathcal{G}^s$ based on the class-wise text prompts outlined in Section \ref{multiclip}. In Eq. \ref{eq:2}, $(\mathbf{I, \mathbf{Y}})$ denotes a sketch or a photo along with the associated one-hot label vector: $\mathbf{I} \in \{\mathcal{P}^s, \mathcal{S}^s\}$, and $\mathbf{Y} = [y^1, \cdots, y^{|\mathcal{C}^s|}]$.
\vspace{-0.2cm}

\begin{equation}
    \mathcal{L}_{\text{class}} = \underset{\scriptsize\begin{array}{c}
    \mathcal{B}_{v}, \mathcal{B}_{t}, \mathcal{B}_{vt}, \\
    {\theta}, {\phi}
    \end{array}}{\min}  \; \underset{(\mathbf{I},\mathbf{Y}) \in \mathcal{G}^s}{\sum}  - \sum_{c=1}^{|\mathcal{C}_{s}|} y^c \log(p(y^c|\mathbf{I}))
\label{eq:2}
\end{equation}

We compute $p(y^{c'}|\mathbf{I})$ as follows, where $\text{Prompt}_{y^{c'}} = $ \texttt{sketch/photo of a} $[CLS_{y^{c'}}]$ represents input text sentence, and $\tau$ is a hyper-parameter.
\vspace{-.2cm}
\begin{equation}
p(y^{c'}|\mathbf{I}) = \frac{ \exp(\cos(\mathcal{F}_{v}(\mathbf{I}), \mathcal{F}_{t}(\text{Prompt}_{y^{c'}}))/\tau)}{\sum_{c=1}^{|\mathcal{C}_{s}|}\ \exp(\cos(\mathcal{F}_{v}(\mathbf{I}), \mathcal{F}_{t}(\text{Prompt}_{y^c}))/\tau)}
\label{prob}
\end{equation}
\\
\noindent \textbf{$\mathcal{L}_{cjs}$: Proposed conditional cross-modal jigsaw loss:} 
Given $(r, r^+, r^-)$ and $y^{\text{perm}}$, $\mathcal{F}_{js}$ follows the approach outlined in \cite{noroozi2016unsupervised} to address the jigsaw task. This involves predicting the permutation index corresponding to $y^{\text{perm}}$ in a list $\mathbf{\Pi}^{1 \times |\mathcal{Y}^{\text{perm}}|}$ that indexes all possible permutations in $\mathcal{Y}^{\text{perm}}$, where $\mathds{1}_{\mathbf{\Pi}(y^{\text{perm}})}$ represents the one-hot encoding for the index of $y^{\text{perm}}$.

The loss function $\mathcal{L}_{cjs}$ serves two objectives. Firstly, utilizing $r$ and $\mathds{1}_{\mathbf{\Pi}(y^{\text{perm}})}$, we aim to minimize the cross-entropy loss $\mathcal{L}_{ce}$, enabling $\mathcal{F}_{js}$ to learn the jigsaw task effectively. Additionally, we introduce a hinge-loss objective to constrain $r^+$ to yield a lower $\mathcal{L}_{ce}(\mathcal{F}_{js}(r^+), \mathds{1}_{\mathbf{\Pi}(y^{\text{perm}})})$ compared to $\mathcal{L}_{ce}(\mathcal{F}_{js}(r^-), \mathds{1}_{\mathbf{\Pi}(y^{\text{perm}})})$. 
\vspace{-0.2cm}
\begin{equation}
    \mathcal{L}_{cjs} = \underset{\scriptsize\begin{array}{c}
    \mathcal{B}_{v}, \mathcal{B}_{t}, \mathcal{B}_{vt}, \mathcal{F}_{js}, \\
    {\theta}, {\phi}
    \end{array}}{\min} \; \underset{(s_a, s'_a, p_a^+, p_a^-) \in \mathcal{G}^s}{\sum} [\mathcal{L}_{ce}(\mathcal{F}_{js}(r), \mathds{1}_{\mathbf{\Pi}(y^{\text{perm}})}) + \mathcal{L}_{margin}]
\label{decoder_loss}
\end{equation}

where $\mathcal{L}_{margin}$ is defined as,

\begin{equation}
    \centering
    \mathcal{L}_{margin} = [\mathcal{L}_{ce}(\mathcal{F}_{js}(r^+), \mathds{1}_{\mathbf{\Pi}(y^{\text{perm}})}) - \mathcal{L}_{ce}(\mathcal{F}_{js}(r^-), \mathds{1}_{\mathbf{\Pi}(y^{\text{perm}})})]_+
\end{equation}

\noindent \textbf{$\mathcal{L}_{total}$: Total loss for training:}
The total loss can be denoted as follows with $\alpha$ and $\beta$ denoting the relative loss weights.\vspace{-0.1cm}
\begin{equation}
    \mathcal{L}_{total} = \mathcal{L}_{triplet} + \alpha * \mathcal{L}_{class} + \beta * \mathcal{L}_{cjs}
\label{total_loss_fg}
\end{equation}

\noindent \textbf{Inference:} During inference, we eliminate the dependence on test class labels from $\mathcal{C}^u$. For generating the tokens $\mathbf{V}^{\texttt{ms}}$, where only the class embeddings are required, we propose to utilize class names from $\mathcal{C}^s$, mirroring the approach used during training, where all the class names are used to define the inputs to $\mathcal{B}_{vt}$. This strategy aims to map test classes into the discriminative space defined by the training classes, effectively combating model bias. The retrieval process of photos, in response to sketch queries, both from $\mathcal{G}^u$, is executed within $\mathcal{F}_v$'s visual embedding space using a nearest-neighbor ranking mechanism. 

\section{Experiments and Results}
\vspace{-0.1cm}
\noindent {\textbf{- Datasets:}} We evaluate \textsc{SpLIP} on three benchmark datasets for categorical (G)ZS-SBIR: Sketchy-Ext \cite{sempcyc, caae}, TU-Berlin-Ext \cite{tuberlinextend}, and QuickDraw-Ext \cite{doodle}, following the established training-validation protocols \cite{bda, sempcyc}. For FG-ZS-SBIR, which necessitates precise sketch-photo matching, we incorporate the Sketchy dataset \cite{sketchy}. Further details, including the splits of Sketchy-1-Ext \cite{sempcyc} \& Sketchy-2-Ext \cite{caae} are described in the \textbf{Supplementary}.

\noindent {\textbf{- Implementation details, training and evaluation protocols:}} 
For $\mathcal{F}_v$, we select the ViT-B/32 backbone of CLIP, while the Transformer-based text encoder is considered for $\mathcal{F}_t$.
Besides, $\mathcal{B}_t$ employs three linear layers to convert $\mathbf{E}_0$ into $m=4$ learnable textual tokens. In contrast, $\mathcal{B}_v$ utilizes a singular linear layer to adapt to the visual dimension, producing four learnable visual tokens for a matching batch size. Meanwhile, $\mathcal{B}_{vt}$ incorporates linear layers and employs a bottleneck architecture consisting of two layers (Linear-ReLU-Linear) for the creation of $n=2$ layer-specific visual tokens. Additionally, $\mathcal{F}_{js}$ is designed with two encoder layers, followed by a classifier, to accurately decode the patch arrangements of $s'_a$ given ($p_a^+$/$p_a^-$/$s_a$).

 The training process spans 60 epochs, initiating with a warm-up learning rate of $0.001$ and utilizing the Adam optimizer \cite{adam} alongside a scheduler. The batch size is configured to $192$ for both Sketchy-Ext and TU-Berlin-Ext datasets, whereas a batch size of $64$ is adopted for QuickDraw-Ext.  $\alpha$ and $\beta$ are fixed through grid search.
\begin{table*}
\caption{Comparison for categorical ZS-SBIR.}
\vspace*{-6mm}
\begin{center}
\scalebox{0.7}{
\begin{tabular}{llccc|cc|cc|cc}
\toprule

&\multicolumn{2}{c}{\multirow{2}{*}{\textbf{Methods}}}&\multicolumn{2}{c}{Sketchy-1-Ext \cite{sempcyc}}&\multicolumn{2}{c}{Sketchy-2-Ext \cite{caae}} &\multicolumn{2}{c}{TU-Berlin-Ext}\cite{tuberlinextend}  &\multicolumn{2}{c}{QuickDraw-Ext}\cite{doodle} \\
\cmidrule(lr){4-5}\cmidrule(lr){6-7}\cmidrule(lr){8-9}\cmidrule(lr){10-11}
 &&&mAP@all &P@100 &mAP@200  &P@200  &mAP@all  &P@100  &mAP@all  &P@200 \\
\midrule
\multirow{11}{*}{\rotatebox{90}{CNN}} \hspace{0.2cm}&CVAE \cite{caae} &ECCV'18 &19.6 &28.4 &22.5  &33.3  &00.5  &00.1  &00.3  &00.3 \\

&CC-DG \cite{ccdg} &CVPR'19 &31.1 &46.8 &- &- &24.7 &39.2 &- &- \\
&Doodle \cite{doodle} &CVPR'19 &- &- &46.1  &37.0  &11.0  &12.1  &07.5  &06.8 \\

&SEM-PCYC \cite{sempcyc} &CVPR'19 &34.9 &46.3 &45.9  &37.0  &29.7 &42.6 &17.7  &18.4 \\

&SAKE \cite{sake} &ICCV'19 &54.7 &69.2 &49.7  &59.8  &47.5  &59.9 &-  &-  \\

&Styleguide \cite{styleguide} &TM'20 &37.6 &48.4 &35.8 &40.0 &25.4 &35.5 &- &- \\

&OCEAN \cite{ocean} &ICME'20 &- &- &-  &-  &33.3  &46.7 &-  &-  \\

&DSN \cite{domainsmoothing} &IJCAI'21 &58.3 &70.4 &- &- &48.4 &59.1 &- &- \\

&TCN \cite{tcn} &TPAMI'21 &61.6 &76.3 &51.6 &60.8 &49.5 &61.6 &14.0 &29.8  \\

&BDA \cite{bda} &NC'22 &43.7 &51.4 &55.6  &45.8  &37.4  &50.4 &15.4  &35.5  \\

&Sketch3T \cite{sketch3t} &CVPR'22 &- &- &57.9  &64.8  &50.7  &67.1 &-  &-  \\

\midrule

\multirow{5}{*}{\rotatebox{90}{ViT}} \hspace{0.2cm}&TVT \cite{tvt} &AAAI'22 &64.8 &79.6 &53.1  &61.8  &48.4  &66.2  &14.9  &29.3 \\

&PSKD \cite{pskd} &ACM MM'22 &68.8 &78.6 &56.0  &64.5  &50.2  &66.2  &15.0  &29.8 \\

&SaA \cite{saa} &- &67.1 &76.2 &53.5 &63.0 &49.5 &60.8 &14.8  &- \\

&ZSE-RN \cite{zse} &CVPR'23 &69.8 &79.7 &52.5  &62.4  &54.2  &65.7  &14.5  &21.6 \\

&ZSE-Ret \cite{zse} &CVPR'23 &73.6 &80.8 &50.4  &60.2  &56.9  &63.7  &14.2  &20.2 \\
\midrule

\multirow{5}{*}{\rotatebox{90}{CLIP}} \hspace{0.2cm}&CLIP-AT \cite{clipforall} &CVPR'23 &- &- &72.3  &72.5  &65.1  &73.2  &20.2  &38.8 \\

&TLT \cite{tlt} &MMTA'23 &77.9 &84.3 &66.1  &73.0  &61.5  &69.5  &27.8  &- \\

&Sherry \cite{adapt} &- &74.1 &83.5 &61.6  &69.5  &54.1  &66.4  &18.0  &29.8 \\

&MARL \cite{marl} &WACV'24 &- &- &69.1  &75.5  &70.5  &77.7  &32.7  &42.5 \\

\cmidrule(lr){2-11}

&\cellcolor{gray!20}\textbf{\textsc{SpLIP}} &\cellcolor{gray!20} &\cellcolor{gray!20}\cellcolor{gray!20}\textbf{80.2}
&\cellcolor{gray!20}\cellcolor{gray!20}\textbf{86.7} &\cellcolor{gray!20}\cellcolor{gray!20}\textbf{76.4} &\cellcolor{gray!20}\textbf{77.3} &\cellcolor{gray!20}\textbf{73.1}&\cellcolor{gray!20}\textbf{78.2} &\cellcolor{gray!20}\textbf{34.2} &\cellcolor{gray!20}\textbf{44.6}\\

\bottomrule
\end{tabular}}
\label{zssbir}
\end{center}
\vspace{-9mm}
\end{table*}
Following the literature \cite{doodle, pskd}, our evaluation for ZS-SBIR considers the top $200$ retrieved photos, where we report the mean Average Precision score (mAP@all) and precision at $200$ (P@200). Aligning with the recent trends, however, we specifically report precision at $100$ (P@100) for the TU-Berlin-Ext dataset and mAP at $200$ (mAP@200) for the Sketchy dataset. For FG-ZS-SBIR, accuracy is evaluated by considering only a single category at a time \cite{pang2020solving}, denoted as Acc@$\mathcal{K}$ for Sketchy. This metric reflects the percentage of sketches that have their true matched photo within the top-$\mathcal{K}$ list, with our focus being on the Top-1 and Top-5 accuracy metrics \cite{clipforall}.
\vspace{-0.2cm}
\subsection{Comparison to the literature} 
\begin{figure*}[ht!]
    \centering
    \includegraphics[scale=0.3]{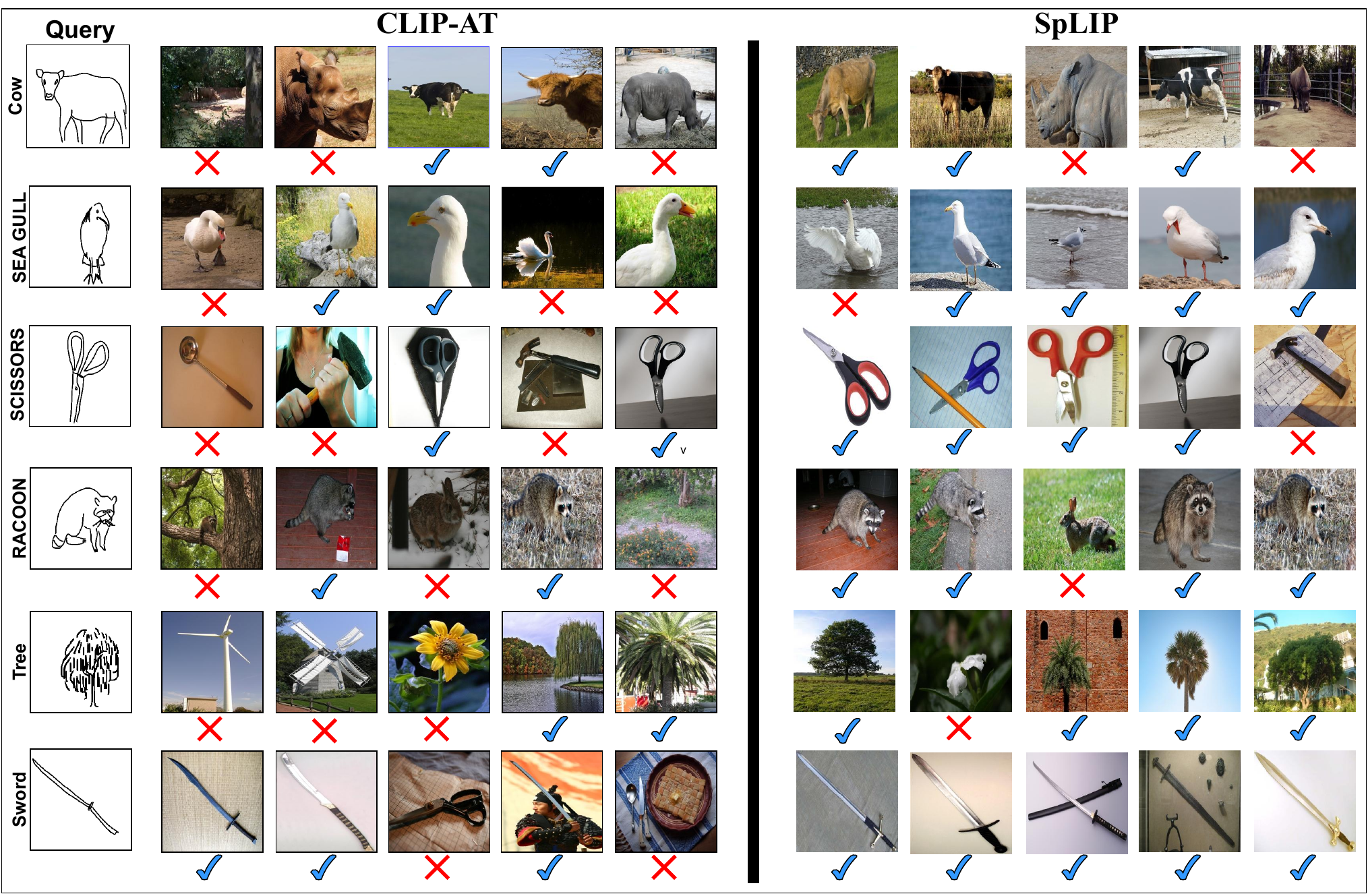}
    \vspace{-.3cm}
    \caption{\small{\textbf{Qualitative comparisons} between \textsc{SpLIP} and \cite{clipforall} in categorical ZS-SBIR. Improved retrieval outcomes for more ambiguous classes can be observed with \textsc{SpLIP}.}}
    \label{retrieved}
    \vspace{-0.8cm}
\end{figure*}

\vspace{-0.2cm}
\noindent {\textbf{- (G)ZS-SBIR and FG-ZS-SBIR:}} Table \ref{zssbir} provides a comparative analysis of \textsc{SpLIP} against methods utilizing ConvNet, ViT, and CLIP backbones across the extended versions of the two splits of the Sketchy, TU-Berlin, and QuickDraw datasets, respectively in categorical ZS-SBIR. Remarkably, \textsc{SpLIP} outperforms all models based on ConvNet and ViT backbones by a substantial margin across all metrics. For instance, on the challenging QuickDraw-Ext dataset, \textsc{SpLIP} surpasses the leading ConvNet-based method, BDA \cite{bda}, by \(18.8\%\) and the premier ViT-based model, PSKD \cite{pskd}, by \(19.2\%\) in the mAP@all metric. In comparison with CLIP-based models, \textsc{SpLIP} achieves an improvement over MARL by \(1.5\%\) in mAP values, with analogous enhancements observed across all datasets, indicating at least a \(2\%\) boost in mAP scores. In Fig. \ref{retrieved}, we show qualitative comparisons between our proposed \textsc{SpLIP} and \cite{clipforall} on ZS-SBIR, highlighting our improved retrieval results for different categories.

Further insights into the performance on more stringent GZS-SBIR tasks are provided in Table \ref{gzssbir}, where \textsc{SpLIP} significantly demonstrates its ability to alleviate model bias towards training classes. It outmatches the second-best approach by \(4.8\%\) on Sketchy-Ext and \(4.1\%\) on TU-Berlin-Ext in mAP scores, unequivocally evidencing enhanced generalization capabilities.
\begin{table}[ht!]
\vspace{-0.4cm}
\begin{minipage}{.5\linewidth}
    \captionof{table}{Comparison for the GZS-SBIR.}
    \vspace{-0.8cm}
    \begin{center}
\scalebox{0.68}{
\begin{tabular}{llccc|cc}
\toprule

&\multicolumn{2}{c}{\multirow{2}{*}{\textbf{Methods}}}&\multicolumn{2}{c}{Sketchy-2-Ext\cite{caae}} &\multicolumn{2}{c}{TU-Berlin-Ext\cite{tuberlinextend}} \\
\cmidrule(lr){4-5}\cmidrule(lr){6-7}
 &&&mAP@200  &P@200  &mAP@all  &P@100 \\
\midrule
\multirow{3}{*}{\rotatebox{90}{CNN}} \hspace{0.2cm}&SEM-PCYC \cite{sempcyc} &CVPR'19 &-  &-  &19.2 &29.8\\

&OCEAN \cite{ocean} &ICME'20 &-  &-  &31.2  &34.1\\

&BDA \cite{bda} &NC'22 &22.6  &33.7  &25.1  &35.7\\

\midrule

\multirow{4}{*}{\rotatebox{90}{ViT}} \hspace{0.2cm}&SaA\cite{saa} &- &- &- &29.0 &38.1 \\

&ZSE-Ret \cite{zse} &CVPR'23 &-  &-  &46.4  &48.5 \\

&ZSE-RN \cite{zse} &CVPR'23 &-  &-  &43.2  &46.0 \\

&STL \cite{stl} &AAAI'23 &63.4 &53.8 &40.2 &49.8 \\

\midrule

\multirow{3}{*}{\rotatebox{90}{CLIP}} \hspace{0.2cm}&CLIP-AT \cite{clipforall} &CVPR'23 &55.6 &62.7 &60.9 &63.8 \\

&MARL \cite{marl} &WACV'24 &62.3 &68.5 & 62.6 &67.8 \\

\cmidrule(lr){2-7}

&\cellcolor{gray!20}\textbf{\textsc{SpLIP}} 
&\cellcolor{gray!20}
&\cellcolor{gray!20}\textbf{68.2} &\cellcolor{gray!20}\textbf{74.5} &\cellcolor{gray!20}\textbf{66.7}&\cellcolor{gray!20}\textbf{70.3}\\

\bottomrule
\end{tabular}}
\label{gzssbir}
\end{center}
\end{minipage}\hspace{10mm}
\begin{minipage}{.4\linewidth}
\vspace{-.56cm}
    \captionof{table}{Comparison for the FG-ZS-SBIR on the Sketchy dataset.}
\vspace{-3mm}
\begin{center}
\scalebox{0.8}{
\begin{tabular}{llccrr}
\toprule

&\multicolumn{1}{l}{\multirow{2}{*}{\textbf{Methods}}}&\multicolumn{2}{c}{Sketchy \cite{sketchy}} \\\cmidrule(lr){3-4}
& &Acc@1  &Acc@5 \\

\midrule
&CrossGrad \cite{crossgrad} &13.40 &34.90\\

&CC-DG \cite{ccdg} &22.60  &49.00\\

&SketchPVT \cite{sketchpvt} &30.24 &51.65 \\

&CLIP-AT \cite{clipforall} &28.68 &62.34 \\

&MARL \cite{marl} &29.96 &58.53 \\

\cmidrule(lr){2-4}

&\cellcolor{gray!20}\textbf{\textsc{SpLIP}} &\cellcolor{gray!20}\cellcolor{gray!20}\textbf{33.45} &\cellcolor{gray!20}\textbf{66.71}\\

\bottomrule
\end{tabular}}
\label{fgzssbir}
\end{center}
\end{minipage}
\vspace{-.75cm}
\end{table}
Table \ref{fgzssbir} showcases the FG-ZS-SBIR performance on the Sketchy dataset, comparing top-1 and top-5 retrieval accuracies. \textsc{SpLIP} attains a top-1 accuracy of \(33.45\%\), leading the subsequent best method by \(3.21\%\). This represents a nearly \(4\%\) improvement over the findings in \cite{clipforall}. Our foundational model, spotlighting multimodal prompting and $\mathcal{L}_{triplet} + \mathcal{L}_{class}$, outperforms the achievements of \cite{clipforall}, highlighting the critical importance of multimodal prompting alone. Additionally, the adoption of $\mathcal{L}_{cjs}$ enhances results by almost \(3.4\%\) in Acc@1.

\begin{wraptable}{r}{0.42\textwidth}
\vspace*{-4mm}
\caption{Comparison of ZS-SBIR across datasets while training with the Sketchy-Ext \cite{caae} dataset and tested on TU-Berlin-Ext and Quickdraw-Ext datasets. $^{*}$ represents the results reproduced by us.}\label{acrossdataset}
\vspace*{-5mm}
\begin{center}
\scalebox{0.65}{
\begin{tabular}{llcc|cc}
\toprule

&\multicolumn{1}{l}{\multirow{2}{*}{\textbf{Methods}}}&\multicolumn{2}{c}{TU-Berlin-Ext} &\multicolumn{2}{c}{Quickdraw-Ext} \\
\cmidrule(lr){3-4}\cmidrule(lr){5-6}
 &&mAP@all  &P@100  &mAP@all  &P@100 \\
\midrule

&CC-DG \cite{ccdg} &30.8 &43.4 &15.6 &22.7 \\

&DSN \cite{domainsmoothing} &35.6 &46.9 &14.9 &17.8 \\

&SAKE \cite{sake} &38.9  &50.6  &17.4  &24.2\\

&ZSE-RN \cite{zse} &47.6  &59.0  &22.8  &33.8 \\


&CLIP-AT$^{*}$ \cite{clipforall} &56.4 &63.1 &30.7 &45.0 \\


&\cellcolor{gray!20}\textbf{\textsc{SpLIP}} &\cellcolor{gray!20}\textbf{70.6} &\cellcolor{gray!20}\textbf{76.0} &\cellcolor{gray!20}\textbf{45.8}&\cellcolor{gray!20}\textbf{58.6}\\

\bottomrule
\end{tabular}}
\end{center}
\vspace{-9mm}
\end{wraptable} 
\noindent {\textbf{- Across dataset ZS-SBIR:}} 
Leveraging on the across-dataset generalizability capabilities of CLIP, we assess the effectiveness of our proposed \textsc{SpLIP} on the ZS-SBIR across dataset setting by \cite{zse}, where the model is trained on the Sketchy-Ext dataset and evaluated on 21 unseen classes of TU-Berlin-Ext and 11 unseen classes of Quickdraw-Ext datasets, respectively. In Table \ref{acrossdataset}, our findings demonstrate that \textsc{SpLIP} outperforms CLIP-AT \cite{clipforall} in terms of mAP@all and P@100 metrics by $14.2\%$ and $12.9\%$ on the TU-Berlin-Ext dataset, and by $15.1\%$ and $13.6\%$ on the Quickdraw-Ext dataset.
\vspace{-4mm}










\vspace*{-3mm}
\subsection{Ablation analysis}
\label{ablation}
\vspace*{-2mm}
Besides the following, ablations on the number of training samples, the ratio of known and unknown classes, etc., are mentioned in the \textbf{Supplementary}.

\noindent {\textbf{- Analysis of the loss components:}} In Table \ref{loss_ablation}, we explore the influence of various loss components as described in Section \ref{loss} on the ZS-SBIR and FG-ZS-SBIR tasks, utilizing the Sketchy dataset for evaluation. Incorporation of $\mathcal{L}_{triplet}$ demonstrates a notable performance improvement, which is further sharply enhanced by $\mathcal{L}_{class}$. Adding $\mathcal{L}_{cjs}$ contributes to an approximate $3-4\%$ increase in results for both tasks. Specifically, we find that both the components of $\mathcal{L}_{cjs}$ show improvements, \ie, using $\mathcal{L}_{class}+\mathcal{L}_{triplet}+\mathcal{L}_{margin}$ improves the results of $\mathcal{L}_{class}+\mathcal{L}_{triplet}$ by $\approx 1-2 \%$, and the use of full $\mathcal{L}_{cjs}$ offers further improvements by $2 \%$ for ZS-SBIR and $\approx 2-3 \%$ for FG-ZS-SBIR.

\noindent {\textbf{- Ablation with learnable blocks of \textsc{SpLIP}:}} In Table \ref{block_ablation}, we elucidate the significance of each learnable module within \textsc{SpLIP}, $\mathcal{B}_v$, $\mathcal{B}_t$, $\mathcal{B}_{vt}$, and LayerNorm parameters, for both ZS-SBIR and FG-ZS-SBIR tasks. Note that excluding one of these blocks means the respective prompt token sharing is disabled.
Consistent with findings from \cite{clipforall}, rendering the parameters of LayerNorm modules learnable results in a performance boost of approximately $2\%$. Conversely, the absence of cross-modal token sharing detrimentally impacts outcomes, whether in one or both directions. Specifically, we note a decline in precision by at least $4\%$ for ZS-SBIR and $3\%$ for FG-ZS-SBIR tasks, underscoring the critical role of token sharing in enhancing model efficacy across both retrieval challenges.

\begin{figure*}[ht!]
    \centering
    \includegraphics[width=0.5\textwidth]{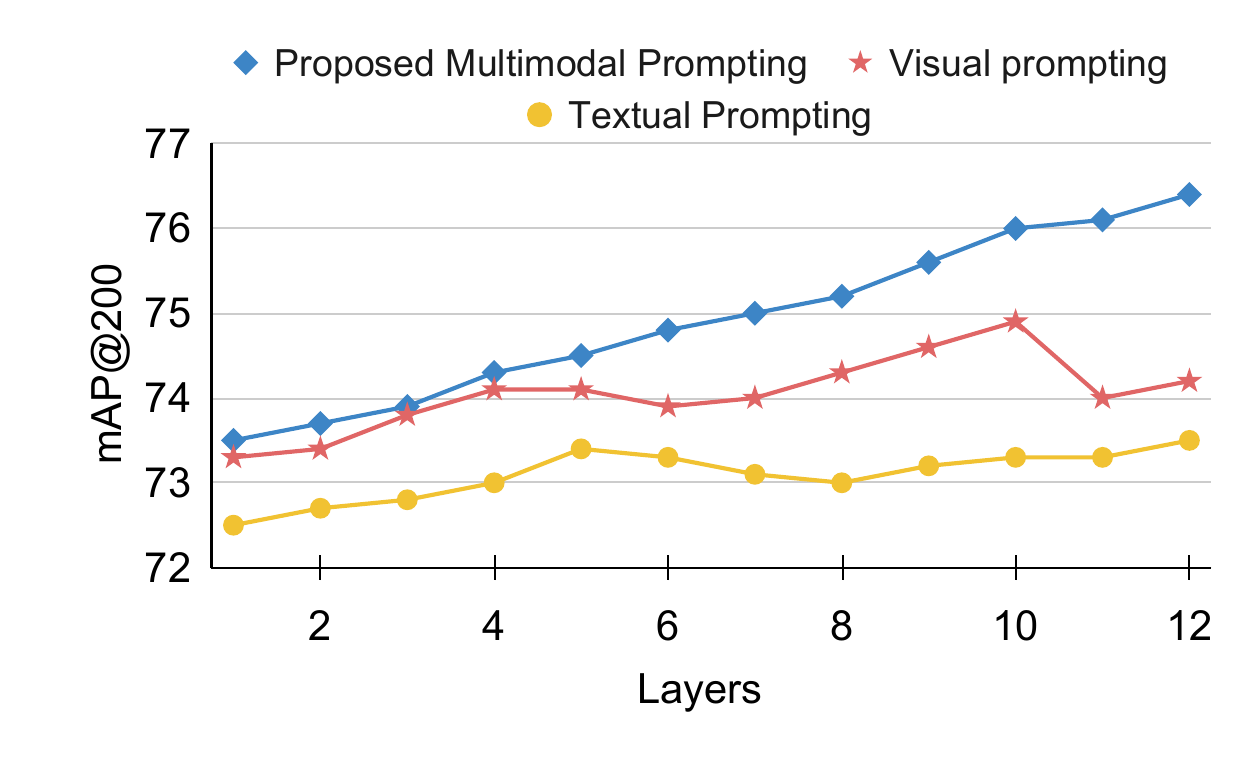}
    \includegraphics[width=0.49\textwidth]{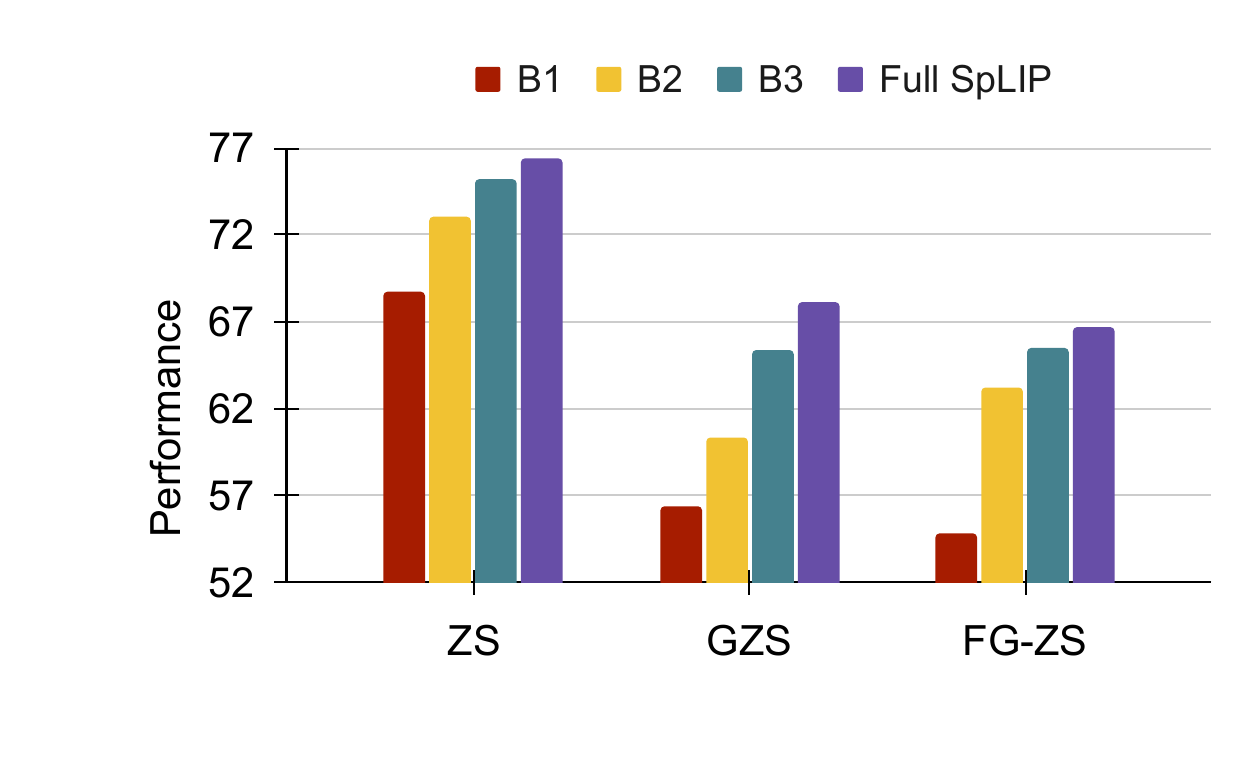}
    \vspace{-0.98cm}
    \caption{\small{\textbf{Left:} Effects of including more layers into our proposed deep prompting. We show the effects of prompting individually for CLIP's vision and textual branches, followed by the proposed multi-modal prompting. \textbf{Right:} B1 refers effects of replacing the multi-modal prompting of ours by that of \cite{maple}, B2 refers to replacing the conditional cross-modal jigsaw solver by the patch permutation-based triplet objective proposed in \cite{clipforall}, and B3 denotes \textsc{SpLIP} with a fixed $\mu$ of $0.2$ for $\mathcal{L}_{triplet}$.}}
    \label{layers}
    \vspace{-0.8cm}
\end{figure*}

\noindent {\textbf{- Shallow to deep prompting in \textsc{SpLIP}:}} Additionally, we assess the effect of employing deep prompting across separate textual and visual branches, alongside our integrated multi-modal approach, as depicted in Figure \ref{layers} (\textbf{Left}). This evaluation reveals a general trend where mAP values ascend with the inclusion of more layers from $\mathcal{F}_v$/$\mathcal{F}_t$ for all three scenarios. Notably, the enhancements achieved through multi-modal prompting distinctly surpass those of the uni-modal counterparts, emphasizing the superior efficacy of cohesively leveraging both textual and visual cues.

\noindent {\textbf{- \textsc{SpLIP} coupled with the multi-modal prompting of \cite{maple}, the fine-grained metric loss of \cite{clipforall}, and with a fixed margin for $\mathcal{L}_{triplet}$:}} Our approach, leveraging extensive bi-directional information sharing between visual and textual modalities, excels beyond current multi-modal prompting methods, including those by \cite{maple}. Contrary to \cite{maple}'s uni-directional and layer-restricted token sharing, our comprehensive integration strategy showcases superior efficacy, as depicted in Fig. \ref{layers} (\textbf{Right}) (\textbf{marked as B1}). 
\begin{table}[ht!]
\vspace*{-6.5mm}
\begin{minipage}{.49\linewidth}
\caption{Ablation of loss terms (Equation \ref{total_loss_fg}) on the Sketchy dataset.}
\vspace*{-9mm}
\begin{center}
\scalebox{0.7}{
\begin{tabular}{llcc|cc}
\toprule

&\multicolumn{1}{l}{\multirow{2}{*}{\textbf{Loss}}}&\multicolumn{2}{c}{ZS-SBIR} &\multicolumn{2}{c}{FG-ZS-SBIR} \\
\cmidrule(lr){3-4}\cmidrule(lr){5-6}
 &&mAP@200 &P@200 &Acc@1 &Acc@5 \\
\midrule

&$\mathcal{L}_{class}$ &57.5 &58.1 &17.23 &39.41 \\

&$\mathcal{L}_{triplet}$ &71.6 &72.7 &26.54 &59.92 \\

&$\mathcal{L}_{class} + \mathcal{L}_{triplet}$ & 73.1 &73.9 &30.07 &62.95  \\

&$\mathcal{L}_{class} + \mathcal{L}_{triplet} + \mathcal{L}_{margin}$ & 74.5 &75.1 &31.23 &63.54 \\

&$\mathcal{L}_{class} + \mathcal{L}_{triplet} + \mathcal{L}_{cjs}$ &\textbf{76.4} &\textbf{77.3} &\textbf{33.45} &\textbf{66.71} \\

\bottomrule
\end{tabular}}
\label{loss_ablation}
\end{center}
\end{minipage}\hspace{5mm}
\begin{minipage}{.45\linewidth}
\caption{Ablation of learnable blocks of \textsc{SpLIP} on Sketchy dataset.}
\vspace{-8mm}
\begin{center}
\scalebox{0.67}{
\begin{tabular}{llcc|cc}
\toprule

&\multicolumn{1}{l}{\multirow{2}{*}{\textbf{Method}}}&\multicolumn{2}{c}{ZS-SBIR} &\multicolumn{2}{c}{FG-ZS-SBIR} \\
\cmidrule(lr){3-4}\cmidrule(lr){5-6}
 &&mAP@200  &P@200 &Acc@1 &Acc@5 \\
\midrule

&w/o LayerNorm &74.2 &74.9 &31.24 &64.14 \\

&w/o $\mathcal{B}_{v}$ &71.9 &73.6 &29.76 &62.58 \\

&w/o $\mathcal{B}_{t}$ &72.8 &73.5 &30.41 &63.39 \\

&w/o $\mathcal{B}_{vt}$ &70.5 &72.0 &28.92 &62.04  \\

&w/o $(\mathcal{B}_{v} + \mathcal{B}_{vt})$ &68.8 &69.7 &26.54 &59.92 \\

&w/o $(\mathcal{B}_{v} + \mathcal{B}_{t} + \mathcal{B}_{vt})$ &62.5 &68.7 &28.49 &59.35 \\

\cmidrule(lr){2-6}

&\cellcolor{gray!20}\textbf{\textsc{SpLIP} (Ours)}
&\cellcolor{gray!20}\textbf{76.4} &\cellcolor{gray!20}\textbf{77.3} &\cellcolor{gray!20}\textbf{33.45}&\cellcolor{gray!20}\textbf{66.71}\\

\bottomrule
\end{tabular}}
\label{block_ablation}
\end{center}
\end{minipage}
\vspace*{-7mm}
\end{table}

We introduced an experiment to underscore the significance of matching patch-permuted sketches with unshuffled photos, thus capturing essential patch-level alignments for better modality congruence. Replacing $\mathcal{L}_{cjs}$ and $\mathcal{F}_{js}$ with \cite{clipforall}'s patch shuffling metric, where a positive pair of sketch-photo shares identical patch permutations and a negative pair does not, our method markedly outperforms the \textsc{SpLIP} combined with \cite{clipforall} across SBIR tasks, highlighting $\mathcal{F}_{js}$'s robustness in \textsc{SpLIP} (Fig. \ref{layers} (\textbf{Right})) (\textbf{marked as B2}).

Opting for a dynamic $\mu$ value in $\mathcal{L}_{triplet}$, over a static $\mu = 0.2$, significantly boosts performance across all tasks, demonstrating our method's nuanced adaptability and effectiveness (Fig. \ref{layers} (\textbf{Right})) (\textbf{marked as B3}).

\vspace{-0.75cm}

\vspace*{-2mm}
\section{Takeaways}
\vspace*{-0.2cm}
This paper unveils \textsc{SpLIP}, a novel CLIP-based framework tailored for ZS, GZS, and FG-ZS SBIR tasks. Our methodology is distinguished by three innovations: a multi-modal prompt learning strategy enhancing cross-modal knowledge sharing and embedding learning within text-sketch-photo triads; the use of CLIP's textual class-name embeddings to dynamically adjust margins for the sketch-photo triplet loss; and a unique conditional cross-modal jigsaw task designed to fine-tune patch-level sketch-photo associations, by repurposing the stand-alone jigsaw task with a metric objective. \textsc{SpLIP}'s efficacy is validated across various benchmarks, consistently showcasing its dominance in all tasks. Our future work will extend to broader vision-language domains, enriching our understanding of visual semantics. The authors sincerely acknowledge the tremendous support from AWL Inc, Japan.


%
%
\bibliographystyle{splncs04}
\bibliography{main}

\clearpage
\appendix

\section{Contents of the supplementary materials}
In this supplementary document, we present detailed information and further experimental results, including:
\begin{itemize}
\item [1.] \textbf{Dataset description and splits for zero-shot SBIR settings}: In Section \ref{data}, we provide the detailed descriptions and splits of the datasets used in our proposed work. 
\item [2.] \textbf{List of notations used in proposed \textsc{SpLIP:}} In Table \ref{notations}, we list the important variables used in \textsc{SpLIP} together with their significance.
\item [3.] \textbf{Ablation with different number of tokens:} Table \ref{token_ablation} signifies the ablation with varying number of tokens produced by $\mathcal{B}_t$, $\mathcal{B}_v$ and $\mathcal{B}_{vt}$.
\item [4.] \textbf{Ablation with varying the hard prompts:} In Table \ref{hardprompt}, we use different hard prompts for the generation of tokens from $\mathcal{B}_v$.
\item [5.] \textbf{Another cross-dataset ZS-SBIR experiment}: We show results of another cross-dataset ZS-SBIR experiment in Table \ref{crossdata}.

\item [6.] \textbf{Ablation on varying the number of training samples and seen classes:} They are reported in Fig. \ref{trainingdata}-\ref{seenclasses}.

\item [7.] \textbf{Analysis of the alignment of the photo and sketch domains:} We compare the same against \cite{clipforall}. We show the t-SNE plots for both the visual domains, as produced by \cite{clipforall} and \textsc{SpLIP} (Fig. \ref{tsne_sketch}-\ref{tsne_photo}), as well as show the domain distances produced by both the methods, in terms of the Fr\'echet distance \cite{dowson1982frechet} (Table \ref{frechet}).

\item [8.] \textbf{Further analysis on the jigsaw task:} We compare the performance of \textsc{SpLIP} on the jigsaw tasks applied on all ($s_a, p_a^+, p_a^-$) sampled from the same visual domain, \eg all from photo or all from sketch and the proposed cross-modal jigsaw. The results are reported in Table \ref{jigsaw}.

\item [9.] \textbf{Further qualitative results:} They are shown  in Fig. \ref{finegrained}.

\end{itemize}

\section{Dataset descriptions}\label{data}
In this work, we evaluate the \textsc{SpLIP} method's effectiveness across leading ZS-SBIR datasets, including Sketchy-Ext, TU-Berlin-Ext, and QuickDraw-Ext. 

\noindent \textbf{Sketchy-Ext} \cite{sempcyc, caae}: This dataset enlarges the original Sketchy collection \cite{sketchy} to 73,002 images across 125 categories, with an average of 604 sketches and 584 images per category. For zero-shot analysis, two distinct splits are employed. In split-1, i.e., Sketchy-1-Ext \cite{sempcyc}, 25 categories are randomly chosen for testing, leaving the remaining 100 for training. On the other hand, Sketchy-2-Ext \cite{caae} isolates 21 classes not present in the ImageNet \cite{imagenet} dataset for testing, with the other 104 classes designated for training.

\noindent \textbf{TU-Berlin-Ext} \cite{tuberlinextend}: An augmentation of the original TU-Berlin \cite{tuberlin} dataset, which included 20,000 sketches over 250 categories. This extension incorporates 204,489 natural images \cite{sketchnet}, averaging 787 images per category, though with notable class-wise image imbalances. Adhering to the partitioning protocol from \cite{sketchhash}, we allocate 30 classes for testing and 220 for training, ensuring each test class has a minimum of 400 images to mitigate the imbalance.

\noindent \textbf{QuickDraw-Ext} \cite{doodle}: A large-scale dataset designed for ZS-SBIR, extending the Google QuickDraw dataset \cite{quickdraw} with detailed photographs. It includes 110 categories, featuring 330,000 sketches (3,000 per category) and 204,000 Flickr-sourced images, each tagged appropriately. The partitioning strategy from \cite{doodle} is used, segregating 30 test classes not found in ImageNet, with the remaining 80 classes utilized for training.

\section{List of important variables and their significance}

\begin{table}[ht!]
    \centering
    \caption{Summarizing the variables used.}
  \scalebox{0.7}{  \begin{tabular}{c|c}\midrule
        \textbf{Notations} & \textbf{Description} \\\hline
         $\mathcal{F}_v$, $\mathcal{F}_t$& Frozen CLIP's visual and text encoders\\
         $\mathcal{F}_{js}$&Transformer-based jigsaw-solver network\\
         $\mathcal{L}$&Number of encoder layers in $\mathcal{F}_v$ and $\mathcal{F}_t$\\
         $s_a$, $s_a'$&Anchor Sketch, Permuted Anchor Sketch\\
        $p_a^+$, $p_a^-$&Positive Photo, Negative Photo, given the anchor sketch\\
        $\mathcal{B}_t$, $\mathbf{T}$&Vision-guided deep textual prompting block, and the obtained tokens\\
        $\mathcal{B}_v$, $\mathbf{V}^{\texttt{tg}}$&Text-guided deep visual prompting block, and the respective semantic domain knowledge defined tokens\\
         $\mathcal{B}_{vt}$, $\mathbf{V}^{\texttt{ms}}$& Vision-text conjunction block, along with the respective tokens\\
         $\mathcal{M}$, $d_t$& Number of embedding tokens and dimension of text tokens\\
         $\mathcal{J}$, $\mathbf{W}_0$& Number of words in input texts and corresponding word embeddings\\
         $\mathbf{E}_0$, $\mathcal{N}$&Initial patch embeddings, number of patch embeddings\\
         $\mathbb{1}$&One-hot label vector\\
         $\theta$, $\phi$& LayerNorm parameters\\ 
         $\mathcal{L}_{ce}$&Cross-entropy loss\\
         $\mathcal{L}_{class}$&Text-image classification loss\\
         $\mathcal{L}_{triplet}$&Cross-visual modality triplet loss\\
         $\mathcal{L}_{margin}$&Margin loss proposed within $\mathcal{L}_{cjs}$\\
         $\mathcal{L}_{cjs}$&Conditional cross-modal jigsaw loss\\
         
         \midrule
    \end{tabular}}
    
    \label{notations}
\end{table}

\section{Ablation with different number of tokens produced by $\mathcal{B}_t$, $\mathcal{B}_v$ and $\mathcal{B}_{vt}$}

We conducted a comprehensive analysis of our \textsc{SpLIP} framework by adjusting the number of tokens generated from the vision-guided deep textual prompting block ($\mathcal{B}_t$), the text-guided deep visual prompting block ($\mathcal{B}_v$), and the vision-text conjunction block ($\mathcal{B}_{vt}$). In Table \ref{token_ablation}, the configuration of token settings is denoted in the order: ($\mathcal{B}_t$, $\mathcal{B}_v$, $\mathcal{B}_{vt}$). We undertook ablation studies across three SBIR tasks: ZS-SBIR, GZS-SBIR, and FG-ZS-SBIR, to gauge the impact of token variation on performance.

The findings from these ablation studies highlight a trend where an increase in the number of tokens correlates with improved performance. Specifically, the configuration with tokens set to (4, 4, 2) achieved superior results for FG-ZS-SBIR tasks on the Sketchy-Ext dataset. Moreover, it was observed that employing a higher number of prompts tends to diminish the raw feature representation, affecting the overall performance of the system. This observation underscores the balance required between the number of prompts and the preservation of feature quality for optimal performance.

\begin{table}[ht!]
\vspace*{-2mm}
\caption{Ablation of number of tokens chosen for all types of prompting. (.,.,.) signifies number of tokens produced by $\mathcal{B}_{t}$, $\mathcal{B}_{v}$, $\mathcal{B}_{vt}$ respectively.}
\begin{center}
\scalebox{1.0}{
\begin{tabular}{cccc|cc|cc}
\toprule

&\multicolumn{1}{c}{\multirow{2}{*}{\textbf{Tokens}}}&\multicolumn{2}{c}{ZS-SBIR}&\multicolumn{2}{c}{GZS-SBIR} &\multicolumn{2}{c}{FG-ZS-SBIR} \\
\cmidrule(lr){3-4}\cmidrule(lr){5-6}\cmidrule(lr){7-8}
 &($\mathcal{B}_t, \mathcal{B}_v, \mathcal{B}_{vt}$)&mAP@200 &P@200 &mAP@200 &P@200 &Acc@1 &Acc@5 \\
\midrule

&(1,1,1) &71.6 &72.4 &60.4 &65.8 &26.54 &56.71 \\

&(1,4,1) &73.1 &73.7 &64.5 &69.8 &29.63 &60.92 \\

&(4,1,1) &74.5 &74.9 &66.0 &71.9 &31.85 &63.07 \\

&(4,4,1) & 76.2 &77.0 &\textbf{68.4} &\textbf{74.6} &33.34 &66.59  \\

&(4,4,2) &\textbf{76.4} &\textbf{77.3} &68.2 &74.5 &\textbf{33.45} &\textbf{66.71} \\

&(4,4,3) &75.9 &77.1 &68.1 &74.3 &33.24 &66.43 \\

&(4,4,4) & 75.3 &76.4 &67.4 &73.6 &32.85 &65.96  \\
\bottomrule
\end{tabular}}
\label{token_ablation}
\end{center}
\end{table}

\section{Ablation with varying the hard textual prompts}
In Table \ref{hardprompt}, we examine the impact of employing hard prompting strategies on the performance of the three considered SBIR tasks.

\begin{table}[ht!]
\vspace*{-2mm}
\caption{Ablation of varying hard textual prompts.}
\vspace*{-2mm}
\begin{center}
\scalebox{1.0}{
\begin{tabular}{cccc|cc|cc}
\toprule

&\multicolumn{1}{c}{\multirow{2}{*}{\textbf{Hard Prompt}}}&\multicolumn{2}{c}{ZS-SBIR}&\multicolumn{2}{c}{GZS-SBIR} &\multicolumn{2}{c}{FG-ZS-SBIR} \\
\cmidrule(lr){3-4}\cmidrule(lr){5-6}\cmidrule(lr){7-8}
 &&mAP@200 &P@200 &mAP@200 &P@200 &Acc@1 &Acc@5 \\
\midrule

&\textit{photo/sketch of a} [CLS] &\textbf{76.4} &\textbf{77.3} &\textbf{68.4} &\textbf{74.6} &31.07 &64.36 \\

&\textit{an image of a} [CLS] &75.6 &76.3 &67.2 &73.6 &33.14 &65.97 \\

&\textit{visual representation of} [CLS] &75.2 & 76.1 & 67.0 & 73.3 &\textbf{33.45} &\textbf{66.71} \\

\bottomrule
\end{tabular}}
\label{hardprompt}
\end{center}
\end{table}

\section{More cross-dataset ZS-SBIR results}
Pl. see Table \ref{crossdata}.

\begin{table}[ht!]
\vspace*{-4mm}
\caption{Comparison of ZS-SBIR across datasets while training with the TU-Berlin-Ext dataset and tested on Sketchy-Ext and Quickdraw-Ext datasets. $^{*}$ represents the results reproduced by us.}\label{acrossdataset}
\vspace*{-5mm}
\begin{center}
\scalebox{1.0}{
\begin{tabular}{llcc|cc}
\toprule

&\multicolumn{1}{l}{\multirow{2}{*}{\textbf{Methods}}}&\multicolumn{2}{c}{Sketchy-Ext} &\multicolumn{2}{c}{Quickdraw-Ext} \\
\cmidrule(lr){3-4}\cmidrule(lr){5-6}
 &&mAP@all  &P@100  &mAP@all  &P@100 \\
\midrule

&CC-DG \cite{ccdg} &62.4 &69.3 &23.1 &29.6 \\

&DSN \cite{domainsmoothing} &61.3 &65.4 &21.8 &24.6 \\

&SAKE \cite{sake} &62.6  &70.1  &23.5  &31.8\\

&ZSE-RN \cite{zse} &74.6  &81.6  &27.3  &37.6 \\


&CLIP-AT$^{*}$ \cite{clipforall} &78.2 &85.9 &33.6 &43.5\\


&\cellcolor{gray!20}\textbf{\textsc{SpLIP}} &\cellcolor{gray!20}\textbf{84.1} &\cellcolor{gray!20}\textbf{90.3} &\cellcolor{gray!20}\textbf{37.0}&\cellcolor{gray!20}\textbf{47.4}\\

\bottomrule
\end{tabular}}\label{crossdata}
\end{center}
\end{table} 

\section{Ablation on the number of training samples, and the ratio between seen and unseen classes}

\begin{figure*}[ht!]
    \centering
    \textbf{\includegraphics[width=0.32\textwidth]{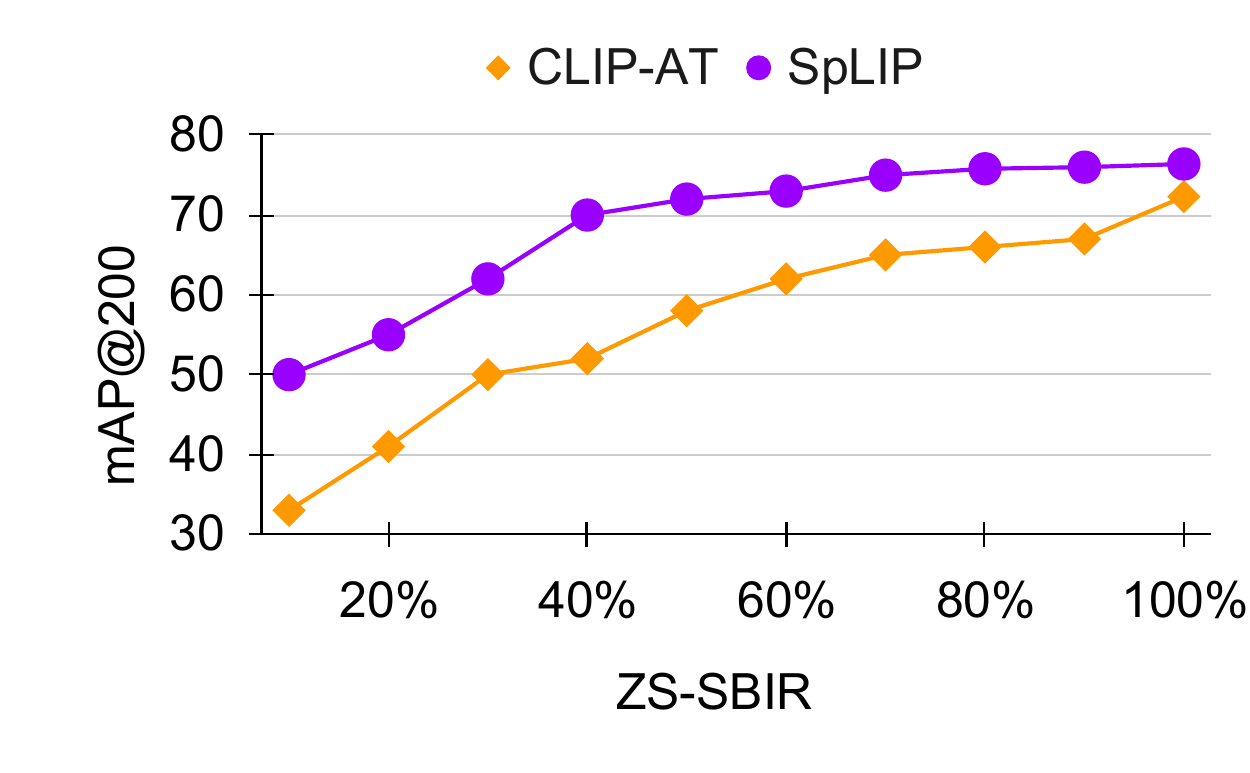}
    \includegraphics[width=0.32\textwidth]{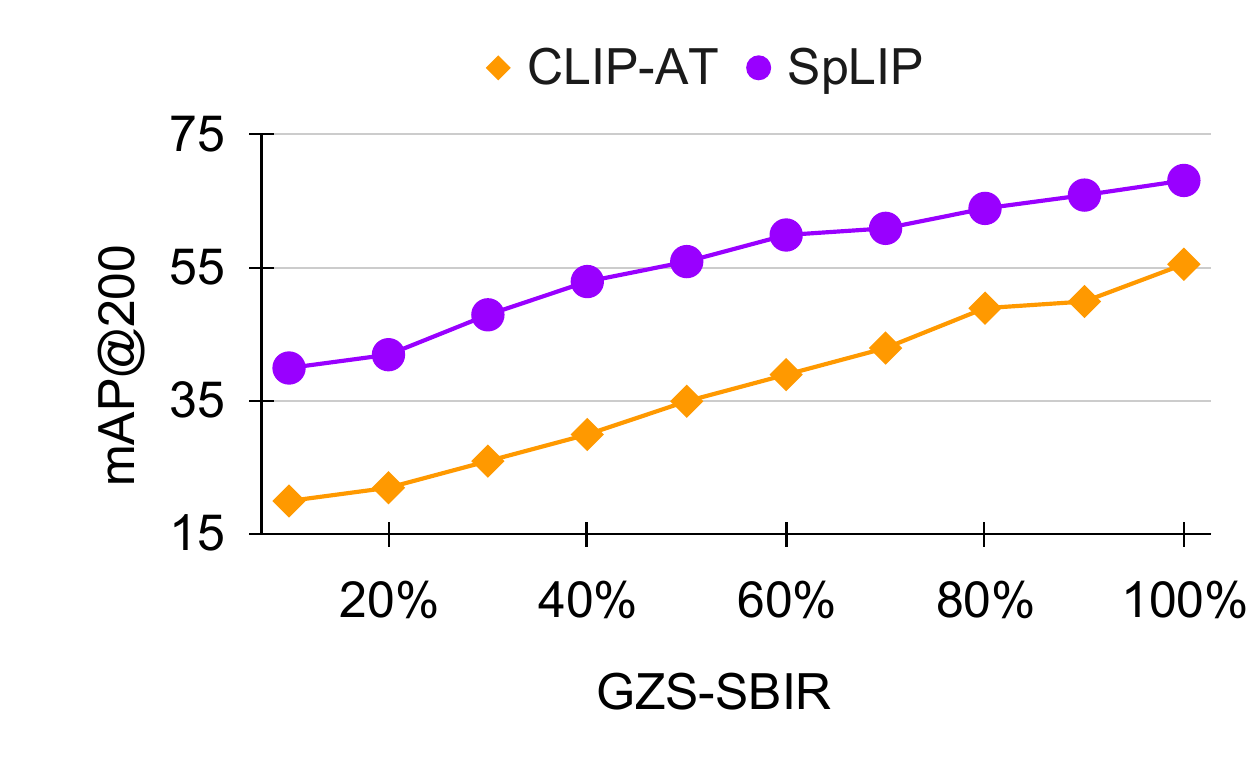}
    \includegraphics[width=0.32\textwidth]{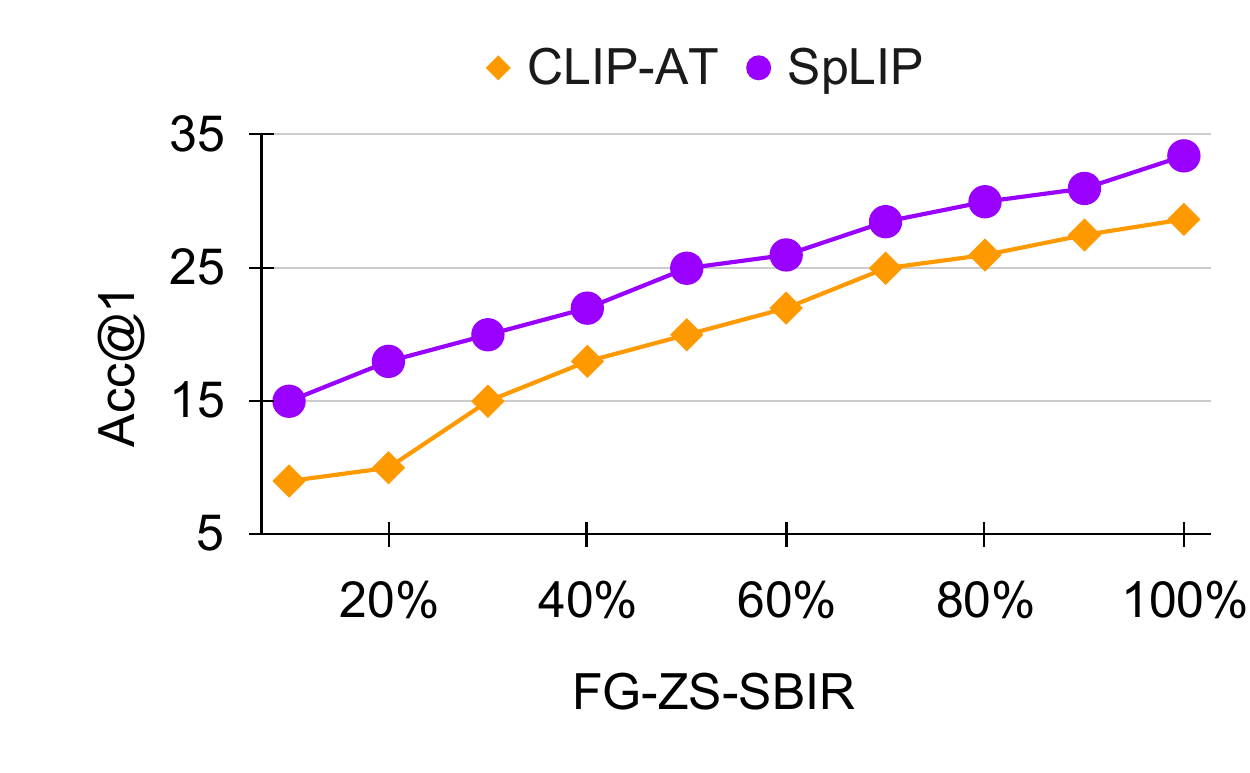}}
    \caption{Performance of \cite{clipforall} and \textsc{SpLIP} with varying training data size for Sketchy-Ext dataset.}
    \label{trainingdata}
\end{figure*}

\begin{figure*}[ht!]
    \centering
    \textbf{\includegraphics[width=0.32\textwidth]{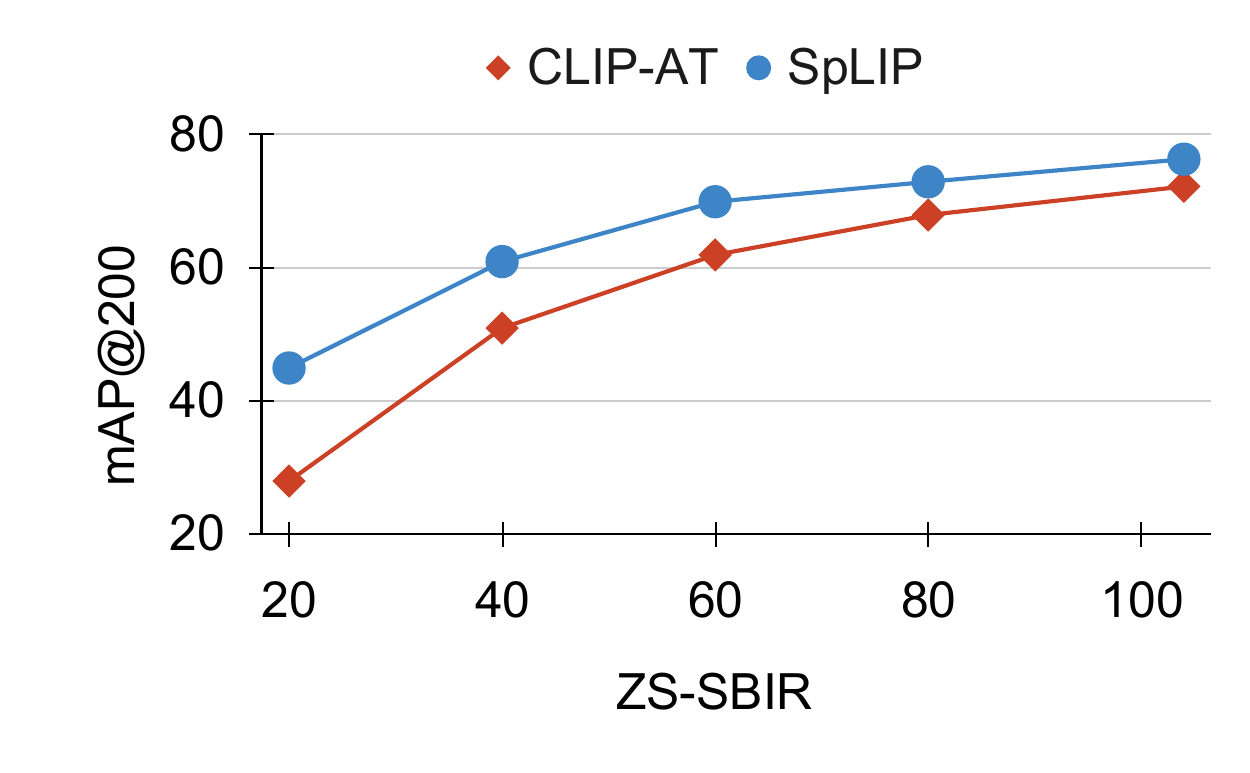}
    \includegraphics[width=0.32\textwidth]{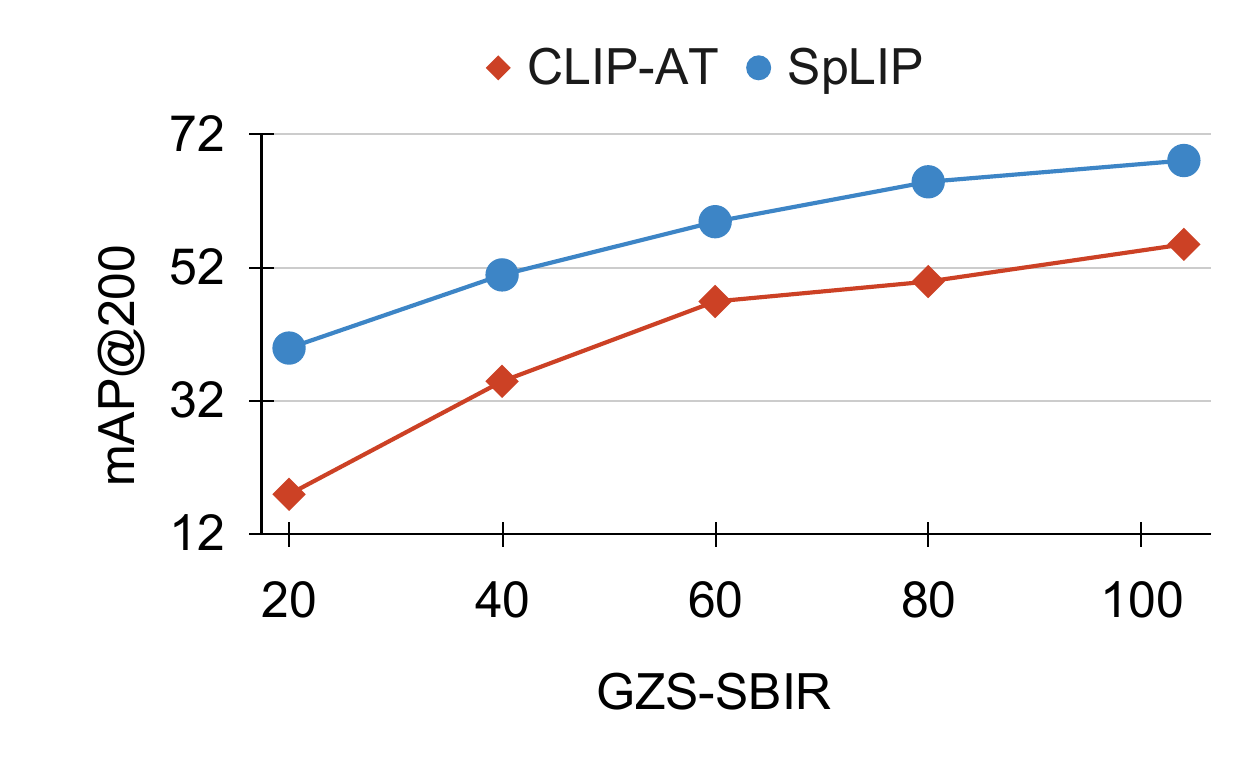}
    \includegraphics[width=0.32\textwidth]{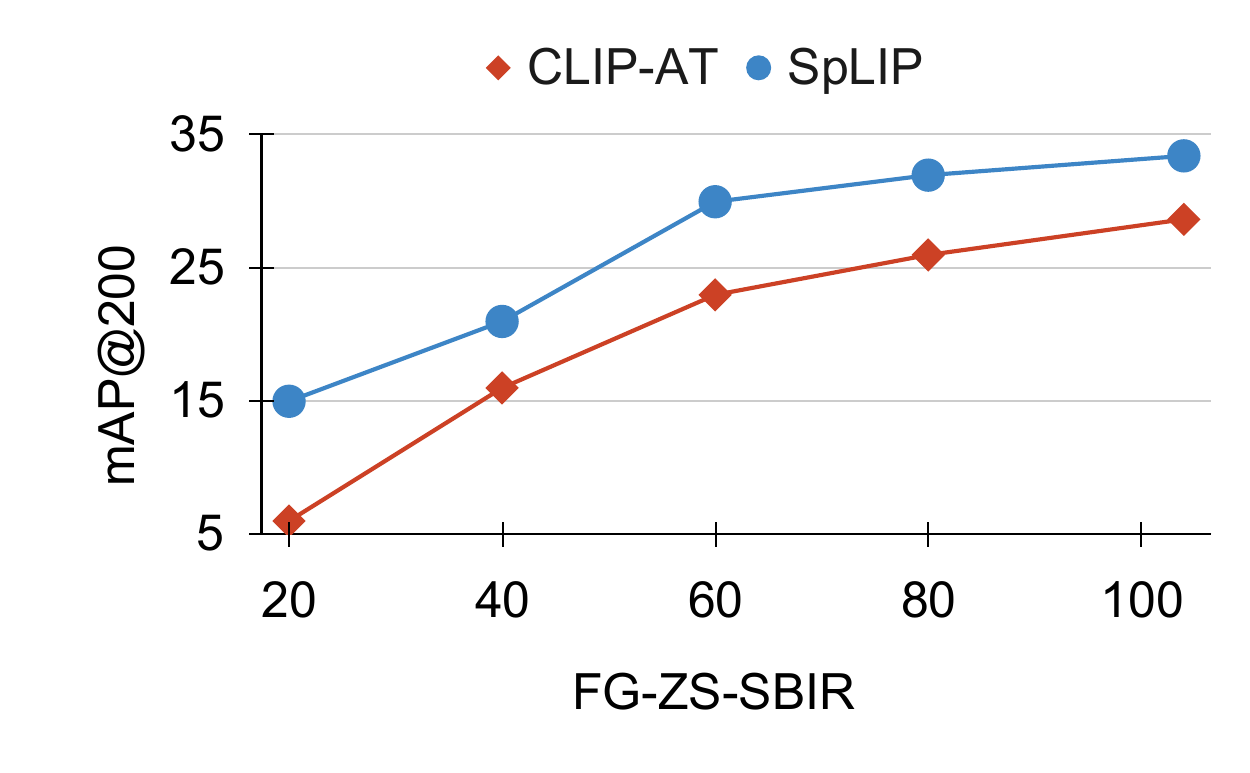}}
    \caption{Performance of \cite{clipforall} and \textsc{SpLIP} with varying number of seen classes while training for Sketchy-Ext dataset.}
    \label{seenclasses}
\end{figure*}

Pl. refer to Fig. \ref{trainingdata} and Fig. \ref{seenclasses} for the same.

\section{Alignment of the sketch and photo domains}

\begin{figure*}[ht!]
    \centering
    \textbf{\includegraphics[width=0.5\textwidth]{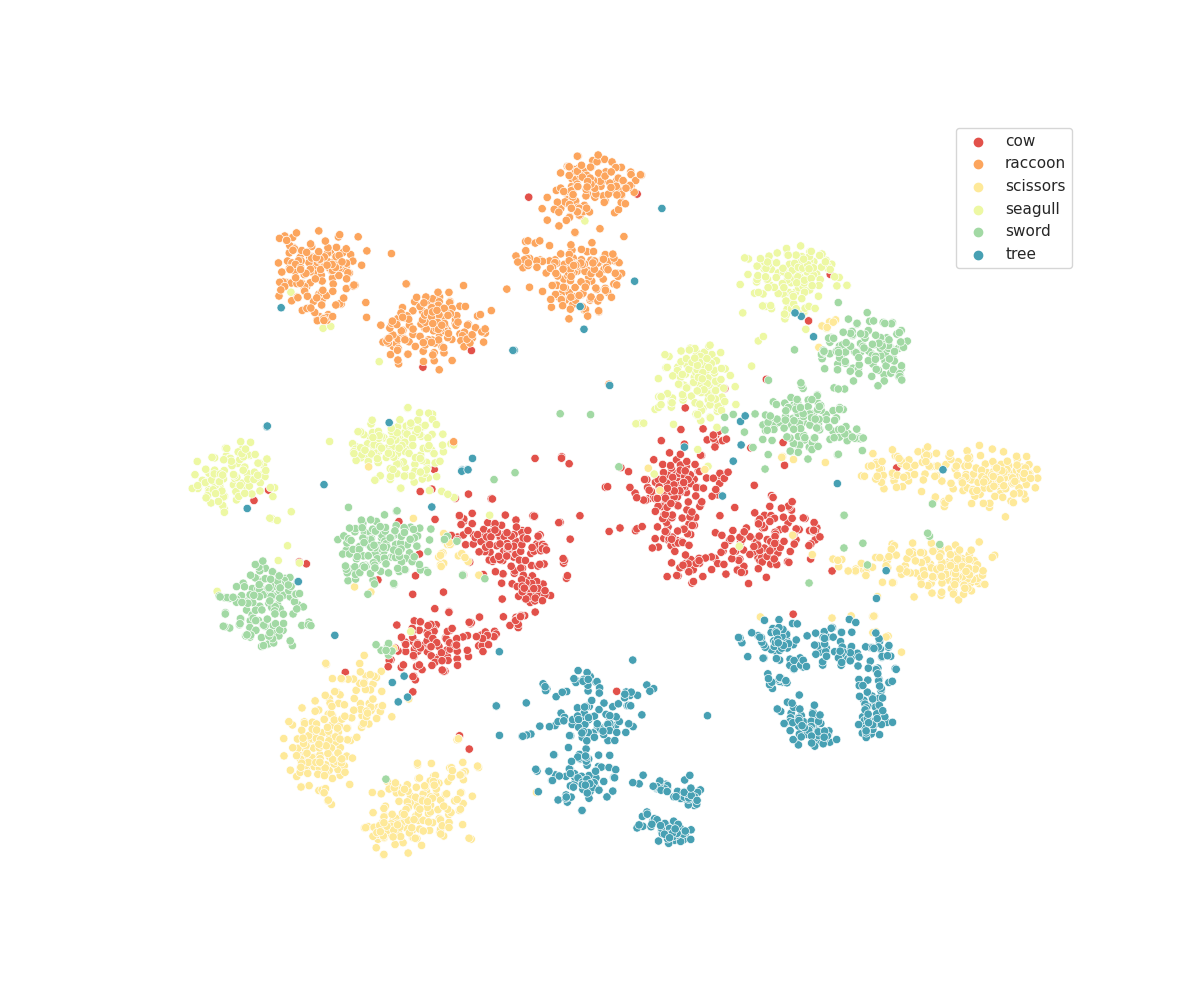}
    \includegraphics[width=0.49\textwidth]{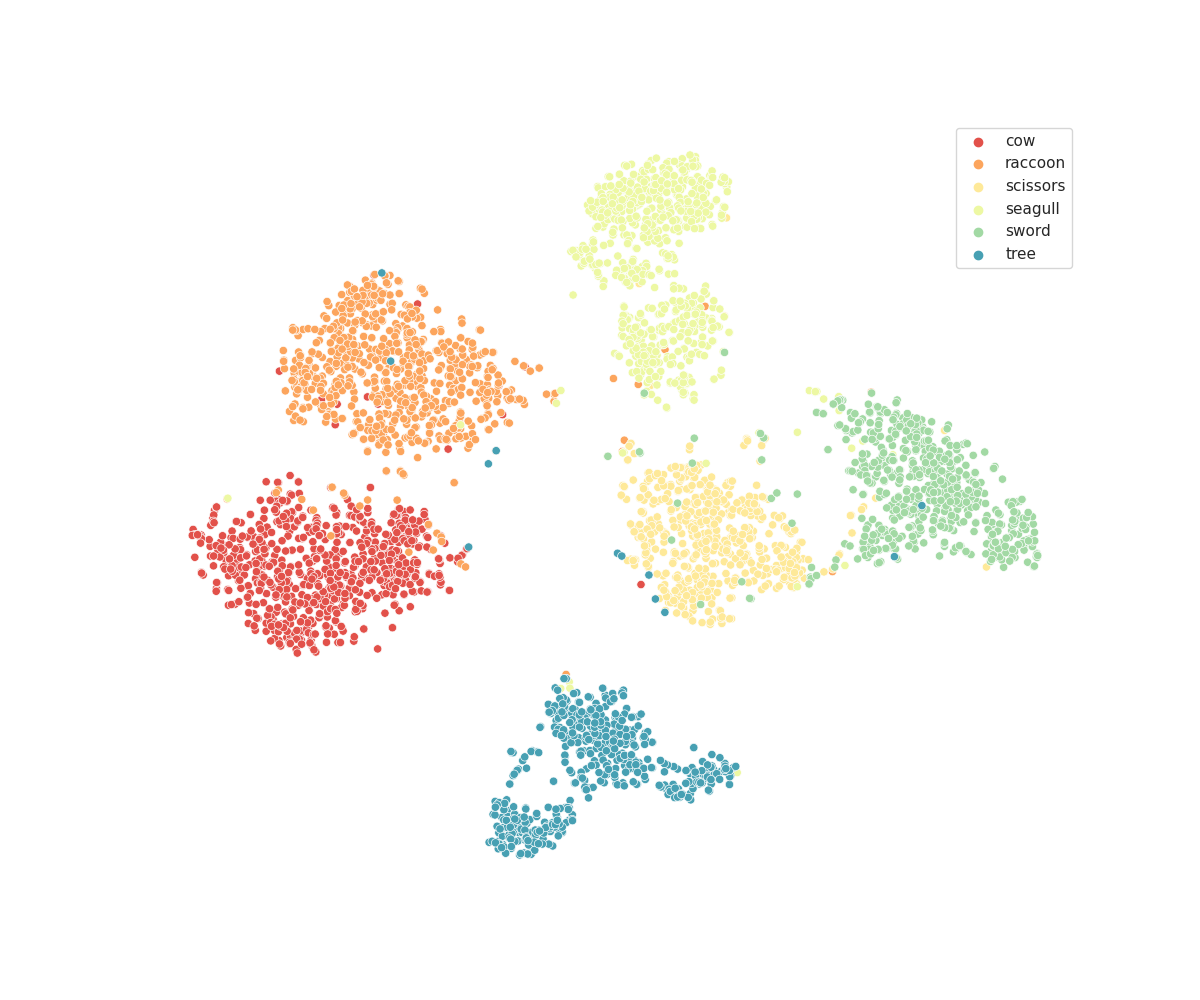}}
    \vspace{-0.7cm}
    \caption{t-sne of sketch domain, by \cite{clipforall} and \textsc{SpLIP}.}
    \label{tsne_sketch}
\end{figure*}

\begin{figure*}[ht!]
    \centering
    \textbf{\includegraphics[width=0.5\textwidth]{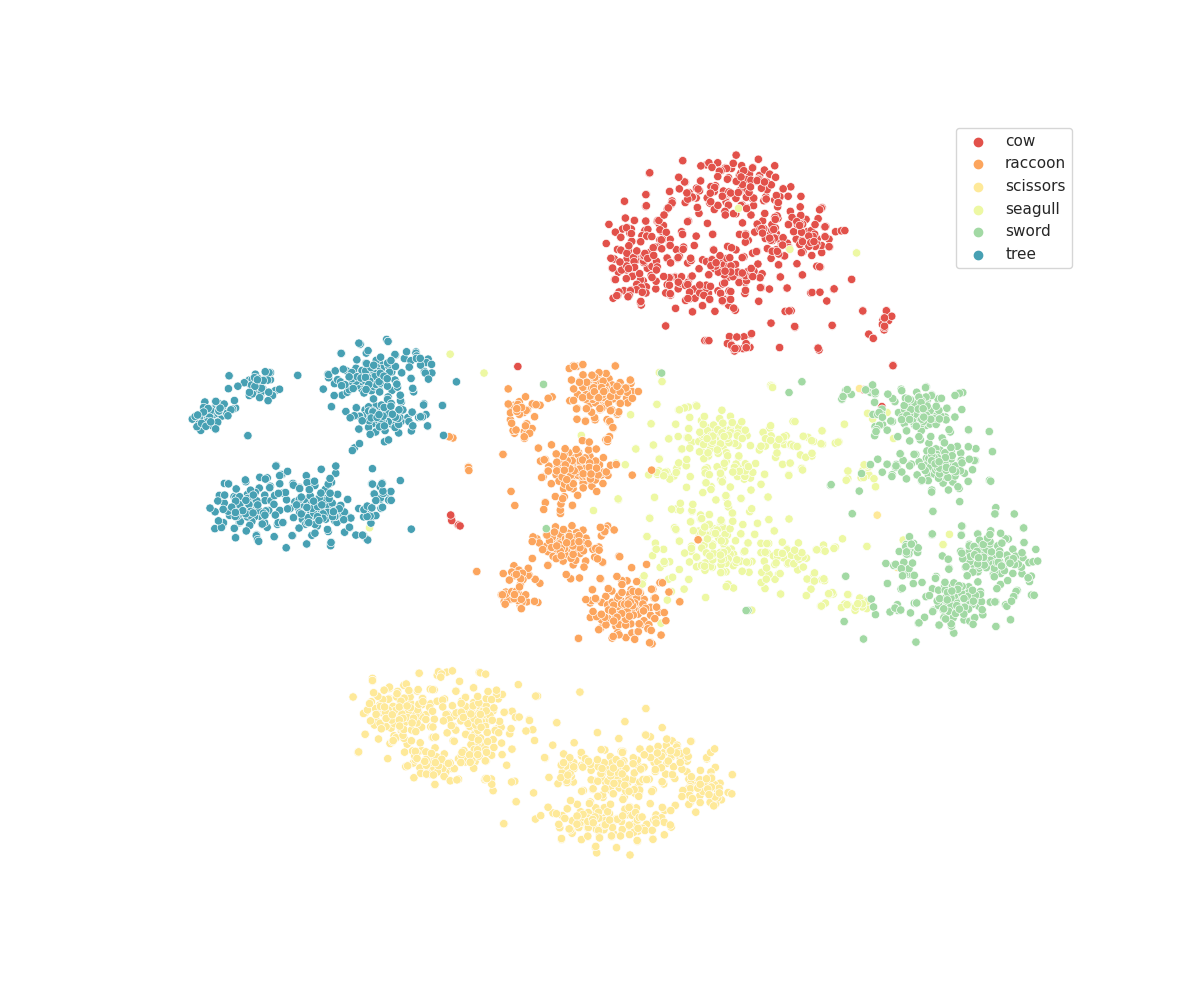}
    \includegraphics[width=0.49\textwidth]{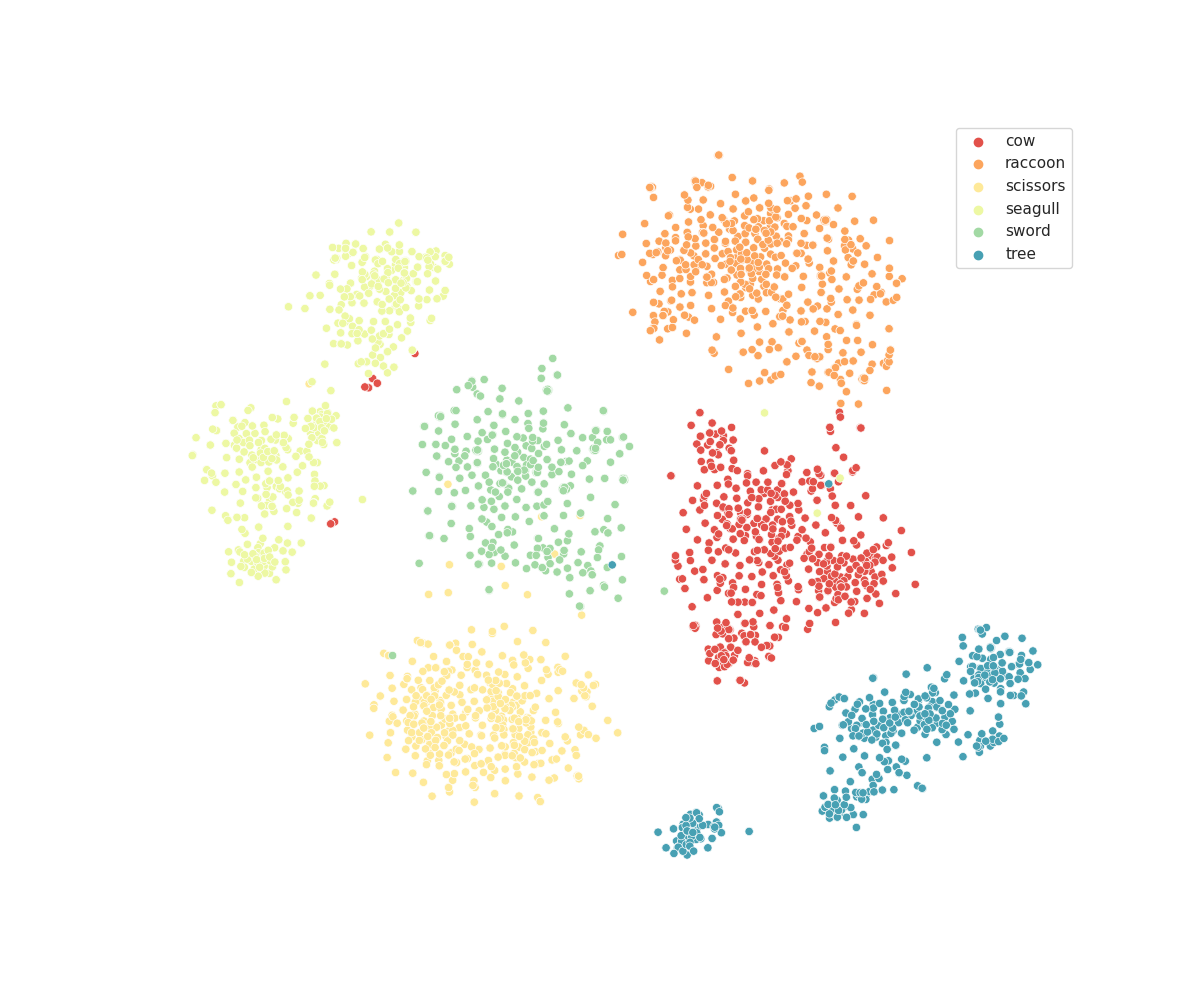}}
    \vspace{-0.7cm}
    \caption{t-sne of photo domain, by \cite{clipforall} and \textsc{SpLIP}.}
    \label{tsne_photo}
\end{figure*}

We present t-SNE visualizations for both modalities, generated by \cite{clipforall} and \textsc{SpLIP}, in Figures \ref{tsne_sketch} and \ref{tsne_photo}, respectively. These visualizations demonstrate that \textsc{SpLIP} achieves superior class separability. Additionally, we assess the domain alignment by comparing the Fre'chet distance between the modalities in the embedding space, as detailed in Table \ref{frechet}. Our results indicate that \textsc{SpLIP} achieves a lower Fre'chet distance, signifying enhanced alignment between domains.

\begin{table}[ht!]
\vspace*{-2mm}
\caption{Fr\'echet distance between sketch and photo in the embedding space on Sketchy-ext dataset.}
\vspace*{-3mm}
\begin{center}
\scalebox{1.0}{
\begin{tabular}{cc|c|c}
\toprule

Method &ZS-SBIR &GZS-SBIR &FG-ZS-SBIR \\
\midrule

CLIP-AT \cite{clipforall} &0.528 &0.617 &0.725 \\

\textsc{SpLIP} &\textbf{0.443} &\textbf{0.485} &\textbf{0.634} \\

\bottomrule
\end{tabular}}
\label{frechet}
\end{center}
\end{table}

\section{Ablation on the jigsaw task}
In Table \ref{jigsaw}, we examine the performance sensitivity of \textsc{SpLIP} for the cross-modal conditional jigsaw task. This assessment includes a comparison against a baseline variant where the anchor, along with its positive and negative examples, originate from a single modality, specifically either sketches or photos. Our findings reveal that cross-modal conditioning outperforms the uni-modal approach by a margin of 3-4\% across all datasets examined.

\begin{table}[ht!]
\vspace*{-2mm}
\caption{Ablation on the jigsaw task. We compare the cross-modal triplets against within-modality triplet selection.}
\vspace*{-2mm}
\begin{center}
\scalebox{1.0}{
\begin{tabular}{cccc|cc|cc}
\toprule

&\multicolumn{1}{c}{\multirow{2}{*}{\textbf{$\mathcal{L}_{cjs}$}}}&\multicolumn{2}{c}{ZS-SBIR}&\multicolumn{2}{c}{GZS-SBIR} &\multicolumn{2}{c}{FG-ZS-SBIR} \\
\cmidrule(lr){3-4}\cmidrule(lr){5-6}\cmidrule(lr){7-8}
 &&mAP@200 &P@200 &mAP@200 &P@200 &Acc@1 &Acc@5 \\
\midrule

&uni-modal &73.2 &74.0 &64.9 &71.3 &29.82 &63.05 \\

&cross-modal &\textbf{76.4} &\textbf{77.3} &\textbf{68.4} &\textbf{74.6} &\textbf{33.45} &\textbf{66.71} \\

\bottomrule
\end{tabular}}
\label{jigsaw}
\end{center}
\end{table}

\section{More qualitative results for FG-ZS-SBIR}

\begin{figure}[ht!]
    \centering
    \includegraphics[width=\textwidth]{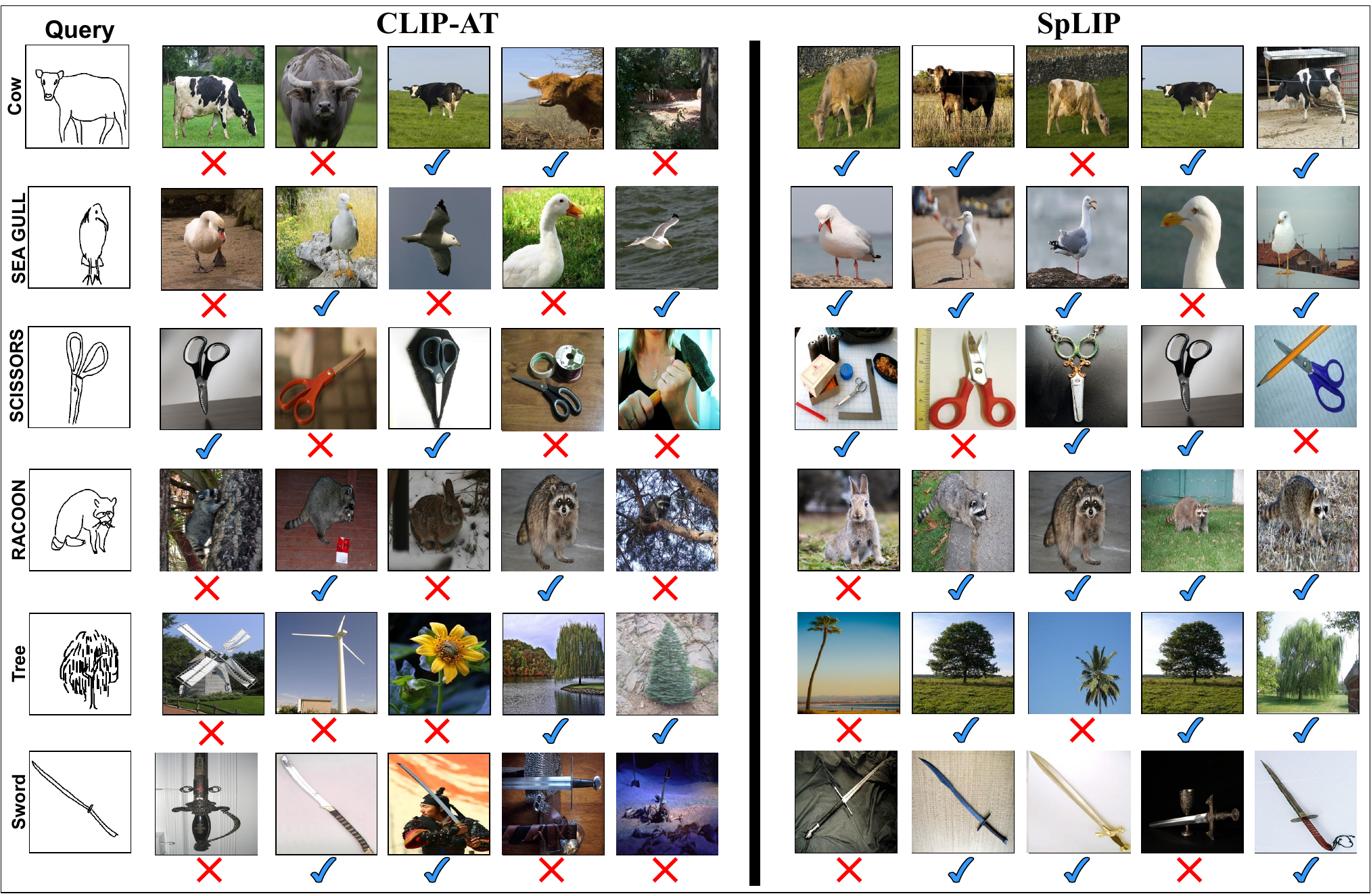}
    \caption{Retrieved photos for sketch query instances for FG-ZS-SBIR.}
    \label{finegrained}
\end{figure}

Pl see Fig. \ref{finegrained}, for results produced by \textsc{SpLIP} (ours) and \cite{clipforall}.

%
%

\end{document}